\documentclass[runningheads]{llncs}
\usepackage[T1]{fontenc}
\usepackage{float}
\usepackage{makecell}
\usepackage{graphicx}
\usepackage{subcaption}
\usepackage{booktabs} 
\usepackage{multirow}
\usepackage{siunitx}
\usepackage{tabularx}
\usepackage{algorithm}
\usepackage{algpseudocode}
\usepackage[leftcaption]{sidecap}
\usepackage{amssymb}
\usepackage{algorithm}
\usepackage{amsmath}
\usepackage{comment}
\usepackage{bm}

\begin{document}

\title{Depth Priors in Removal Neural Radiance Fields}

\author{Zhihao Guo\inst{1} \and
Peng Wang\inst{2} }

\institute{Department of Computing and Mathematics, Manchester Metropolitan University, 
\email{zhihao.guo@stu.mmu.ac.uk, p.wang@mmu.ac.uk}\\}
\maketitle              
\begin{abstract}
Neural Radiance Fields (NeRF) have achieved impressive results in 3D reconstruction and novel view generation. A significant challenge within NeRF involves editing reconstructed 3D scenes, such as object removal, which demands consistency across multiple views and the synthesis of high-quality perspectives. Previous studies have integrated depth priors, typically sourced from LiDAR or sparse depth estimates from COLMAP, to enhance NeRF's performance in object removal. However, these methods are either expensive or time-consuming. This paper proposes a new pipeline that leverages SpinNeRF and monocular depth estimation models like ZoeDepth to enhance NeRF's performance in complex object removal with improved efficiency. A thorough evaluation of COLMAP’s dense depth reconstruction on the KITTI dataset is conducted to demonstrate that COLMAP can be viewed as a cost-effective and scalable alternative for acquiring depth ground truth compared to traditional methods like LiDAR. This serves as the basis for evaluating the performance of monocular depth estimation models to determine the best one for generating depth priors for SpinNeRF. The new pipeline is tested in various scenarios involving 3D reconstruction and object removal, and the results indicate that our pipeline significantly reduces the time required for the acquisition of depth priors for object removal and enhances the fidelity of the synthesized views, suggesting substantial potential for building high-fidelity digital twin systems with increased efficiency in the future.

\end{abstract}
\section{Introduction}
\begin{figure}[btph]
    \centering
    \includegraphics[width=1.0\linewidth]{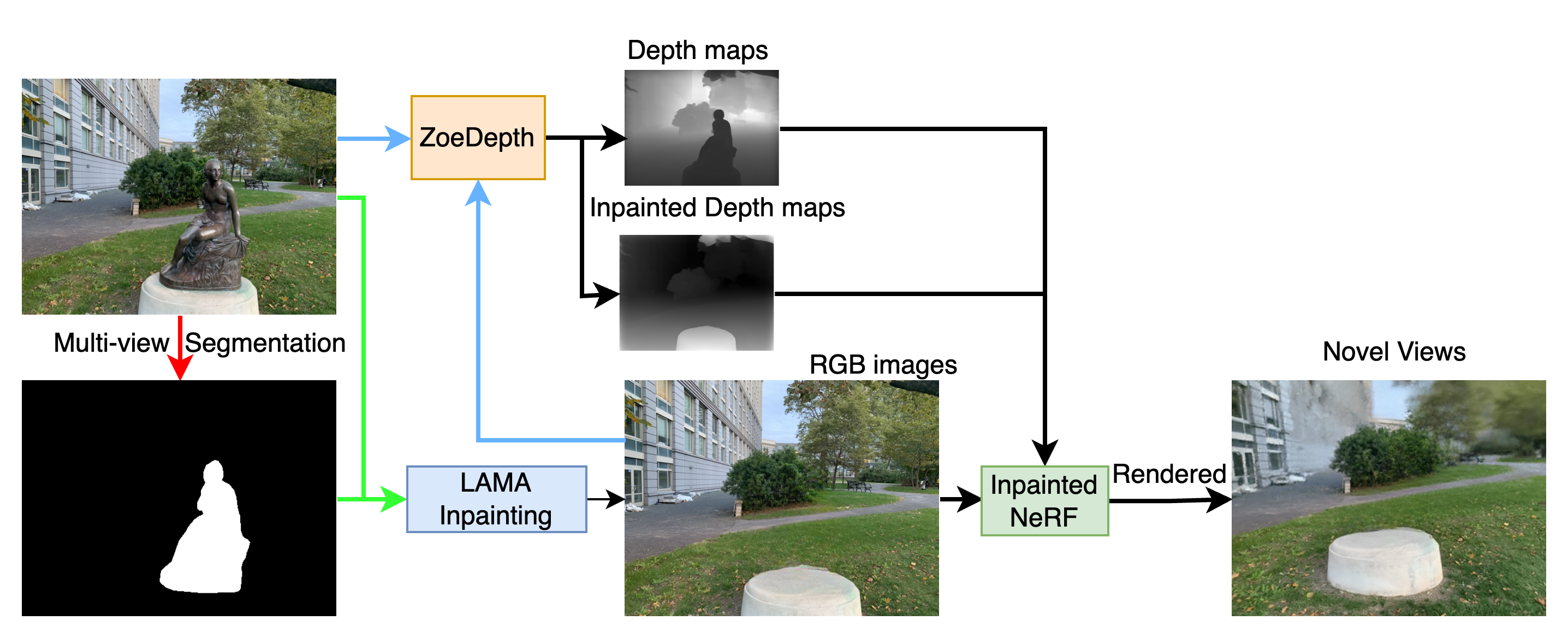}
    \caption{Overview of the proposed object removal pipeline. Starting with sparse views and corresponding masks as inputs, multi-view segmentation (indicated by the red arrow) is employed to generate consistent masks. Depth maps and inpainted depth maps as priors are produced by ZoeDepth (depicted by the blue arrow). The green arrow highlights the multi-view consistent inpainting process, which integrates inpainted depth maps and RGB images into the updated NeRF model to render novel views.}
    \label{fig:spin_zoe}
\end{figure}
Neural Radiance Fields (NeRF) have remarkably transformed the domain of 3D reconstruction and novel view synthesis~\cite{mildenhall2021nerf}. The advancements are promising in revolutionizing sectors from augmented reality to robotic environment perception and modeling. Particularly, it would help to develop high-fidelity digital twins of human-robot collaboration systems in manufacturing settings~\cite{wang2024deep}, facilitating high-consequence decision-making and safe and flexible human-robot collaboration. 
Despite the advancements in NeRF techniques, a persistent challenge within the framework is editing the scenes rendered by NeRF seamlessly, such as removing objects of interest and inpainting the removed area in coherence with the background. This capability is particularly vital in contexts such as human-robot collaborative environments, where the strategic removal of certain obstacles is essential for the planning of safe navigation through environments for robots.

Recent works have addressed this issue by introducing depth priors into NeRF, to supervise the rendering process and improve the rendering quality~\cite{yang2021learning,mirzaei2023spin,weder2023removing}. Typically, depth is provided by LiDAR point clouds or depth completion techniques trained with sparse COLMAP depth, which are either expensive economically or time-consuming to get~\cite{deng2022depth}. Nevertheless, depth priors or depth maps help preserve the fidelity of edited scenes, such as when objects are removed. They play a vital role in maintaining consistency in lighting and shadows, which are key to realistic rendering and accurate perception. Therefore, it is crucial to thoroughly evaluate how different depth priors affect the performance of Removal NeRF.

This paper proposes a novel removal NeRF pipeline which employs SpinNeRF as the foundational reconstruction and removal model. Depth priors generated by various monocular depth estimation models, such as ZoeDepth~\cite{bhat2023zoedepth}, are next integrated into SpinNeRF to enhance its performance and accelerate the speed in object removal tasks, which are evaluated across a range of metrics such as PSNR (Peak Signal-to-Noise Ratio) and SSIM (Structural Similarity Index Measure). It is worth noting that monocular depth estimation models are exploited due to their efficiency and the potential to bring SpinNeRF into real-time systems. 


One challenge facing performance evaluation of monocular depth estimation models is the absence of ground truth depth data. This lack of verifiable depth information necessitates finding substitutes for true depth values where they are unavailable. In response to this issue, an extensive evaluation of COLMAP's capabilities for dense depth reconstruction was performed using various KITTI datasets~\cite{geiger2012we}. The evaluation focused on two key metrics: the precision of the depth reconstruction and the completeness of the depth maps produced. These metrics help assess how accurately and fully COLMAP can infer depth information from monocular images.

Our research findings demonstrate that COLMAP exhibits superior performance in depth estimation, positioning it as a viable substitute for conventional ground truth depth acquisition methods, particularly in scenarios where acquiring true depth information can be prohibitively expensive. Taking depth generated by COLMAP as ground truth, we then evaluated several state-of-the-art monocular depth estimation models on the SpinNeRF dataset~\cite{mirzaei2023spin} to find the optimal model for providing depth priors. ZoeDepth emerges as the leading monocular depth estimation method, delivering high-quality depth priors while concurrently reducing computational overhead. This makes the integration of ZoeDepth with SpinNeRF intuitive to create our proposed pipeline, which substantially improves both the robustness and the quality of the rendered NeRF scenes and the object removal processes. These findings underscore the transformative potential of monocular depth estimation in enhancing NeRF applications. Figure~\ref{fig:spin_zoe} shows an overview of the proposed pipeline.

The contributions of this work include 1). A new pipeline that integrates depth priors from monocular depth estimation models such as ZoeDepth with SpinNeRF enables SpinNeRF to handle complex object removal scenarios and reduce computational overhead. 2). A comprehensive evaluation of COLMAP's dense depth reconstruction capabilities on KITTI dataset helps to establish COLMAP as a viable, cost-effective alternative to conventional ground truth depth data acquisition methods. 3). A systematic evaluation of various monocular depth estimation models, guiding the selection of the optimal depth estimation model to improve reconstruction and object removal performance of NeRF-based methods. 

The remainder of the paper is organised as follows. Some related works are introduced in Section \ref{relatedworks}. Section \ref{Removal NeRF with Monocular Depth Estimation} elaborates on the impacts of depth priors on the removal performance of NeRF, and the methodology integrating SpinNeRF with ZoeDepth is presented. Experiments, discussions, and an ablation study are provided in Section \ref{experiments}, and the paper is concluded in Section \ref{con}.

\section{Related Works}
This section explores NeRF-based 3D reconstruction and depth-enhanced NeRF for object removal. We will also delve into various depth estimation methods to discuss their potential to provide depth priors for the removal NeRF.
\label{relatedworks}
\subsection{Neural Radiance Fields}
\subsubsection{NeRF} 
NeRF~\cite{mildenhall2021nerf} utilise implicit representation for the first time to achieve photo-realistic perspective synthesis, subsequently inferring the 3D structure of the scene. It utilises a limited number of input views to train a neural network to represent a continuous volumetric scene, enabling the generation of new perspectives of the scene after neural network training. To be specific, given the 3D location $\mathbf{p}=(x,y,z)$ of a spatial sampling point (to be rendered) and 2D view direction $\mathbf{d}=(\theta,\phi)$ of the camera, NeRF predicts the color $\mathbf{c}=(r,g,b)$ of the sampling point and the volume density $\sigma$ through a multi-layer perception (MLP) neural network~\cite{riedmiller2014multi} which can be represented as 
\begin{equation}
    F_{\Theta}:(\mathbf{p}, \mathbf{d}) \rightarrow(\mathbf{c}, \sigma),
\end{equation}
\noindent
where $F_{\Theta}$ is the MLP parameterised by $\Theta$.
The color is estimated by volumetric rendering via quadrature, which can be formulated as
\begin{equation}
    C(\mathbf{r})=\int_{t_n}^{t_f} T(t) \sigma\big(\mathbf{r}(t)\big) \mathbf{c}\big(\mathbf{r}(t), \mathbf{d}\big) d t, 
\end{equation}
where $T(t)=\exp \left(-\int_{t_n}^t \sigma\big(\mathbf{r}(s)\big) d s\right)$, $C(\mathbf{r})$ is the sampling pixel value and is calculated by integrating the radiance value $\mathbf{c}$ along the ray $\mathbf{r}(t)=\mathbf{o}+t \mathbf{d}$, in which $\mathbf{o}$ is the camera position, $\mathbf{d}$ is the direction from the camera to the sampled pixel, within near and far bounds $t_n$ and $t_f$, and the function $T(t)$ denotes the accumulated transmittance along each ray from $t_n$ to $t$. While NeRF is renowned for its ability to produce highly detailed and coherent 3D renderings from a relatively sparse set of images, it can be computationally intensive and slow to train and render new perspectives, particularly for complex scenes. In addition, in cases where insufficient input images are used for training, the rendering quality of NeRF can be fairly poor. 

\subsubsection{Depth Supervised NeRF}
Depth priors provide explicit information about the distance between scene surfaces and the camera. Recent advancements have effectively integrated depth priors into NeRF reconstructions to enhance their photorealism and details from sparse inputs. NerfingMVS~\cite{Wei_2021_ICCV} leverages depth from Multi-View Stereo (MVS) to train a depth predictor, significantly enhancing the scene's detail and guiding the NeRF sampling process. This approach effectively uses depth priors to refine the sampling strategy within the NeRF pipeline, thereby improving the model's efficiency and output quality. Similar to NerfingMVS, DONeRF~\cite{neff2021donerf} introduces a depth oracle trained to pinpoint significant surfaces in 3D space, optimising rendering speeds and computational resource allocation. 

Furthermore, Depth-Supervised NeRF (DSNeRF)~\cite{deng2022depth} employs depth priors obtained from 3D point clouds estimated via Structure from Motion (SfM). DSNeRF introduces a novel loss function that aligns the termination of a ray with specific 3D key points, incorporating reprojection errors to gauge uncertainty. This approach ensures that novel view renderings are not only more accurate but also incorporate a measure of confidence in the depth estimation, which is crucial for realistic outputs. Similarly, the study presented in Dense Depth Priors for NeRF~\cite{roessle2022dense} develops a network that enriches sparse depth data from SfM, resulting in dense depth maps that guide both the optimisation process and the scene sampling methodology. By providing depth information where it is typically missing, this approach provides a more robust pipeline for NeRF to produce high-quality reconstructions from limited data. However, one significant challenge for DSNeRF is the substantial time consumption required to generate dense depth information, which can substantially decelerate the process, impeding the practical deployment of this technology in real-time applications where rapid processing is essential.

\subsubsection{Removal NeRF} Removal NeRF is a specialized variant of NeRF designed for editing and removing objects synthesized by NeRF models. This capability is crucial for applications requiring dynamic scene manipulation such as in virtual reality, augmented reality, and potentially digital twins. The techniques developed in this area can be broadly classified into two main categories, each addressing the challenges of object removal with distinct approaches: 

1) Object Compositional NeRF:
This category focuses on learning decomposable components of a scene, enabling selective editing of individual elements. Object-NeRF~\cite{yang2021learning} approaches scene decomposition by learning separate NeRF for different objects and allows for the manipulation of specific scene components without altering the rest of the environment. Despite its effectiveness in isolated edits, maintaining consistency across different viewpoints remains a challenge, particularly when dealing with complex interactions between light and geometry. ObjectSDF~\cite{wu2022object} uses signed distance functions to represent each object within the NeRF pipeline. While ObjectSDF enhances the geometric definition of edited areas, ensuring photorealistic blending and view consistency in diverse lighting conditions continues to pose significant difficulties. 

2) Inpainting-Based Removal: This approach integrates 2D inpainting techniques to reconstruct the scenes following object removal, using depth information to ensure spatial consistency. SPinNeRF~\cite{mirzaei2023spin} leverages the LAMA inpainting method~\cite{suvorov2022resolution} combined with depth priors to enhance the quality of the regions from which objects are removed. This method focuses on generating depth-consistent inpainting to seamlessly blend the edited areas with their surroundings. NeRF-Object Removal~\cite{weder2023removing} builds upon this concept by adapting the inpainting process based on depth cues from the surrounding environment, improving the quality and realism of the final rendered images.

\subsection{Depth Estimation}
\subsubsection{COLMAP Dense Depth} 

COLMAP is an end-to-end multi-view image-based 3D reconstruction technique that is built on top of SfM~\cite{schoenberger2016sfm} and MVS reconstruction~\cite{10.1007/978-3-319-46487-9_31}. In addition, it can also achieve depth estimation, which is required by 3D synthesis models such as removal NeRF. We will briefly go through the pipeline of COLMAP as it is used for depth ground truth generation in this paper for datasets where true depth values are unavailable.

 SfM mainly entails the following procedures: feature extraction $F_i$, feature matching $M_{ij}$, robust model estimation, and bundle adjustment. Feature extraction $F_i$ aims at extracting features such as SIFT from a given image $I_i$. The features from image $I_i$ will be next matched with features from another image(s) $I_j$, and we denote the matching process as $M_{ij}$. The matched features will be used to create a sparse 3D point cloud using triangulation\cite{triggs2000bundle}, which is denoted as
\begin{equation}
P_{ij} = T\Big(M_{ij}, C_i, C_j\Big),
\end{equation}
where \( P_{ij} \) are the 3D points triangulated from matches \( M_{ij} \), and \( C_i \), \( C_j \) are the camera parameters (intrinsic and extrinsic) associated with images \( I_i \) and \( I_j \), respectively. 
The bundle adjustment procedure refines camera parameters and 3D points simultaneously to minimise the re-projection error, and each minimization term individually considers camera parameter $C_i$ and $C_j$, as well as the 3D points $P_{ij}$ to simplify the optimization problem and expedites the computational process, which is denoted as
\begin{equation}
\min _{C_i, C_j, P_{i j}} \sum_{i, j} d\big(Q\left(M_{i j}, C_i, C_j\right), P_{i j}\big),
\end{equation}
where $d(\cdot, \cdot)$ is the Euclidean distance, and \( Q(\cdot,\cdot,\cdot) \) is the projection function that projects 3D points \( P_{ij} \) onto the image plane using camera parameters \( C_i \) and \( C_j \).

\subsubsection{Monocular Depth Estimation}
Depth estimation from monocular images has led to a variety of approaches. Early monocular depth estimation works~\cite{hoiem2007recovering,liu2008sift} relied on handcrafted features, which lack generality to unpredictable and variable conditions in natural scenes. The limitations of these methods have prompted the exploration of deep learning methods~\cite{bhat2023zoedepth,patni2024ecodepth,yang2024depth,Bhat_2021_CVPR,Gasperini_2023_ICCV} in depth estimation.

Some recent works focus on encoder-decoder models~\cite{bhat2023zoedepth,yang2024depth}, for instance, ZoeDepth~\cite{bhat2023zoedepth} introduces a novel architecture combines relative and metric depth, following MiDaS depth estimation framework~\cite{ranftl2020towards} to predict relative depth, and add heads to the encoder-decoder model for metric depth estimation, which is achieved by attaching a metric bins module to the decoder. The final depth of a pixel $i$ is obtained by a linear combination of the bin centers weighted by the corresponding predicted probabilities,
\begin{equation}
   d(i)=\sum_{k=1}^{N_{\text {total }}} p_i(k) c_i(k),
\end{equation}
\noindent
where $d(i)$ is the predicted depth at pixel $i$, $N_{\text {total }}$ is the total number of depth bins, $p_i(k)$ is the probability of the $k-th$ bin for pixel $i$, and $c_i(k)$ is the center of the $k-th$ depth bin. 

DepthAnything~\cite{yang2024depth} is another deep learning-based model depth estimation model. It learns from both unlabeled image datasets $\mathcal{D}^u=\left\{u_i\right\}_{i=1}^N$ and labeled datasets $\mathcal{D}^l=\left\{\left(x_i, d_i\right)\right\}_{i=1}^M$. A teacher model $T$ is first learned from $\mathcal{D}^l$, which is next used to assign pseudo depth labels for $\mathcal{D}^u$. The depth estimation model will then be trained on the combination of labeled datasets and pseudo-unlabeled datasets.

Recently, EcoDepth~\cite{patni2024ecodepth} introduced a new architecture module called Comprehensive Image Detail Embedding (CIDE) and formulated monocular depth estimation as a dense regression problem:
\begin{equation}
\mathbb{P}(\mathbf{y} \mid \mathbf{x}, \mathcal{E})=\mathbb{P}\left(\mathbf{y} \mid \mathbf{z}_0\right) \mathbb{P}\left(\mathbf{z}_0 \mid \mathbf{z}_t, \mathcal{C}\right) \mathbb{P}\left(\mathbf{z}_t \mid \mathbf{x}\right) \mathbb{P}(\mathcal{C} \mid \mathbf{x}),
\end{equation}
\noindent
where $\mathbf{x}$ and $\mathbf{y}$ denote the input image and the output depth, respectively. The terms in the equation are defined as follows: $\mathcal{C}$ represents the semantic embedding derived from the CIDE module.
\begin{equation}
    \mathbb{P}(\mathcal{C} \mid \mathbf{x}) = \mathbb{P}(\mathcal{C} \mid \mathcal{E}) \mathbb{P}(\mathcal{E} \mid \mathbf{x}),
\end{equation}
where $\mathcal{E}$ is the embedding vector from a ViT module~\cite{dosovitskiy2020image}, and $\mathbb{P}(\mathcal{C} \mid \mathcal{E})$ is implemented using downstream modules in CIDE consisting of learnable embeddings.
$\mathbb{P}\left(\mathbf{z}_t \mid \mathbf{x}\right)$ is implemented using the encoder of a variational autoencoder~\cite{takagi2023high}.
$\mathbb{P}\left(\mathbf{z}_0 \mid \mathbf{z}_t, \mathcal{C}\right)$ is implemented using conditional diffusion.
$\mathbb{P}\left(\mathbf{y} \mid \mathbf{z}_0\right)$ is implemented through the depth regressor module, which is a two-layer convolutional neural network.
The final depth map $\mathbf{y}$ is obtained through an upsampling decoder. This decoder includes deconvolution layers and performs a resolution transition, specifically transitioning from a lower resolution of the concatenated feature map back to the original dimensions of the input image.


\section{Removal NeRF with Monocular Depth Priors}
\label{Removal NeRF with Monocular Depth Estimation}
Depth priors are instrumental in delivering essential spatial metrics that substantially influence the performance of NeRF models. These priors provide precise distance measurements between scene objects and the camera, crucial for the accurate reconstruction of 3D scenes. By streamlining the rendering process and facilitating model training, depth priors not only expedite convergence but also minimize computational demands, thereby optimizing the overall efficiency of the NeRF framework. This paper investigates how depth priors affect the object removal performance of NeRF models. To this end, the architecture of SpinNeRF~\cite{mirzaei2023spin} is employed as the base NeRF model. Depth priors obtained by different models such as ZoeDepth are next introduced to SpinNeRF, and the new model's performance on object removal is then evaluated against a series of metrics.

SpinNeRF is a cutting-edge NeRF model designed to enhance the rendering of 3D scenes from sparse viewpoints. Unlike traditional NeRF models, SpinNeRF introduces a unique capability to interpolate and extrapolate views dynamically by learning the underlying structure of a scene through sparse image inputs. To be specific, given a set of RGB images $\mathcal{I}=\left\{I_i\right\}_{i=1}^n$, with the corresponding camera poses $\mathcal{G}=\left\{G_i\right\}_{i=1}^n$ and camera intrinsic matrix $K$ estimated by SfM, and the corresponding removal object mask $\mathcal{M}=\left\{M_i\right\}_{i=1}^n$ provided by users or learned by a different model, SpinNeRF leverages a deep learning framework to encode the volumetric properties of a scene into a neural network, allowing for highly realistic and computationally efficient 3D rendering. During the rendering process, SpinNeRF utilises LAMA\cite{suvorov2022resolution} to achieve image inpainting, and the inpainted images are denoted as $\left\{\widetilde{I}_i\right\}_{i=1}^n$, with $\widetilde{I}_i=\text{INP}\left(\mathcal{I}, \mathcal{M}\right)$ denoting the inpainting process. 

To improve the overall removal and rendering performance, SpinNeRF uses the perceptual loss, LPIPS \cite{zhang2018unreasonable} to optimise the masked regions during rendering, while MSE is still used for unmasked regions,
\begin{equation}
    \mathcal{L}_{\text {LPIPS }}=\frac{1}{|\mathcal{B}|} \sum_{i=1}^{|\mathcal{B}|} \text{LPIPS}\left(\widehat{I}_i, \widetilde{I}_i\right),
\end{equation}
where $\mathcal{B}$ is a subset of the entire dataset of indices from 1 to $|\mathcal{B}|$, $\widehat{I}_i$ is the $i-th$ view rendered using NeRF, $\widetilde{I}_i$ is the $i-th$ view that has been inpainted.

To achieve high-quality image inpainting, depth information on objects of interest is required. In the SpinNeRF pipeline, for each pixel along ray $r$, depth priors $D(r)$  are calculated using
\begin{equation}
    D(r)=\sum_{i=1}^N T_i\Big(1-\exp \left(-\sigma_i \delta_i\right)\Big) t_i.
\end{equation}
Where $N$ is the number of sample points along each ray $r$ from the camera, $t_i$ is the depth of $i-th$ sample point, $\sigma_i$ is the density at the i-th sample, indicating how likely the light is to scatter at this point, $\delta_i$ is the segment length between consecutive samples along the ray, and $T_i$ is the cumulative transmittance to the $i-th$ sample, reflecting the probability that light reaches this point without being obstructed.

Depth priors are then passed to \text{INP} to generate inpainted depth maps $\left\{\widetilde{D}_i\right\}_{i=1}^n$, where $\widetilde{D}_i=\text{INP}(D_i,M_i)$.  The depth maps will be used to supervise the inpainted NeRF,  and the inpainting process is optimised using  the  $\ell_2$ distance between the rendered depths ${D}_i$  and the inpainted depths $\widetilde{D}_i$
\begin{equation}
\mathcal{L}_{\text {depth }}=\frac{1}{R} \sum_{r=1}^R|D(r)-\widetilde{D}(r)|^2,
\end{equation}
Where $\widetilde{D}(r)$ and $D(r)$ are the inpainted depths and rendered depths, $R$ is the total number of rays. 

\section{Experiments and Analysis}
\label{experiments}
\subsection{Dataset}
We have conducted a comprehensive evaluation of dense depth maps reconstructed utilizing COLMAP on a select subset of the KITTI dataset, which includes ground truth depth data. This data is pivotal in verifying the accuracy and practicality of employing COLMAP-generated depth in real-world scenarios, particularly for 3D view synthesis where depth serves as a prior. 
The collection includes 1,048 frames, which are strategically selected to represent a diverse range of conditions and settings.

Our evaluations on the KITTI dataset verifies the hypothesis that dense depth maps produced by COLMAP attain the accuracy required to function as ground truth for applications such as 3D view synthesis. Concurrently, we employed a secondary dataset used by SpinNeRF~\cite{mirzaei2023spin}, comprising 10 distinct scenes accompanied by camera parameters and sparse reconstructed points from SfM. The ground truth depth maps for this dataset were generated through COLMAP's dense reconstruction process.

Building on the affirmative conclusion that dense depth maps generated by COLMAP can be treated as ground truth, we subsequently assess how depth priors derived from alternative sources, such as monocular depth estimation methods, impact object removal in NeRF models. To validate our assumptions, a comprehensive suite of experiments has been conducted.
\subsection{Evaluation Metrics}
\subsubsection{Depth Metrics}
The disparities between depth generated by COLMAP (or real depth from KITTI when available) and monocular depth estimation models are evaluated using different metrics such as  Root Mean Squared Error (RMSE) and the logarithmic scale error~\cite{Godard_2019_ICCV}. Additionally, the widely adopted threshold-based accuracy metric $\delta_1$ is used. This metric specifically measures the proportion of predicted depth values that fall within predefined error thresholds relative to the ground truth. The three metrics are defined as follows
\begin{flalign}
    & \text{RMSE} = \sqrt{{1}/{N} \sum_{i=1}^{N} (D_{\text{gt}_i} - D_{\text{pred}_i})^2}, & \\
    & \delta_1 = \max\left({D_{\text{gt}_i}}/{D_{\text{pred}_i}}, {D_{\text{pred}_i}}/{D_{\text{gt}_i}}\right) < 1.25, & \\
    & \log_{10} \text{Error} = {1}/{N} \sum_{i=1}^{N} |\log_{10}(D_{\text{gt}_i}) - \log_{10}(D_{\text{pred}_i})|, &
\end{flalign}
\noindent
where $N$ is the number of images, and $D_{\text{gt}_i}$ and $D_{\text{pred}_i}$ are the $i-th$ depth ground truth and the corresponding predicted depth, respectively. 

Considering the challenges in calculating the metrics introduced by very sparse ground truth depth values from KITTI, we have designed Algorithm \ref{algo2} to calculate these values.
\begin{algorithm}[btp]
\caption{Dense Depth Evaluation}
\label{algo2}
\begin{algorithmic}[1]
\State \textbf{Inputs:} Predicted depth maps $D_{\text {pred }_i}$ and ground truth depth maps $D_{\text {gt}_i}$
\State \textbf{Outputs:} Mean values of each metric across all frames
\State Initialize metrics lists for RMSE, $\delta_1$, $\delta_2$, $\delta_3$, and $\log _{10}$
\For{each frame in the dataset}
    \State Load ground truth depth map $D_{gt}$
    \State Load predicted depth map $D_{pred}$
    \State Align dimensions of $D_{gt}$ and $D_{pred}$ by padding
    \State Compute scaling factor $s$ and adjust $D_{pred}$ by $s$
    \State Crop to valid region in both $D_{gt}$ and $D_{pred}$
    \State Compute performance metrics for the frame
\EndFor
\end{algorithmic}
\end{algorithm}
\subsubsection{NeRF Metrics}
The widely used metrics for synthesized image quality evaluation such as PSNR and SSIM~\cite{wang2004image,lin2023towards} are used to evaluate the performance of object removal using NeRF with different depth priors. The definitions of PSNR and SSIM are as follows.
\begin{equation}
\text{PSNR} = 20 \cdot \log_{10}\left(\frac{\max_I}{\sqrt{\text{MSE}}}\right),
\end{equation}
where $\max_I$ is the maximum possible pixel value of the image, and $\text{MSE}$ stands for the mean squared error between the original and reconstructed images. 
\begin{equation}
\text{SSIM}(x, y) = \frac{(2\mu_x \mu_y + c_1)(2\sigma_{xy} + c_2)}{(\mu_x^2 + \mu_y^2 + c_1)(\sigma_x^2 + \sigma_y^2 + c_2)},
\end{equation}
where $x$ is the original depth map pixel, $y$ is the corresponding predicted depth map pixel.$\mu_x$ is the average of $x$, $\mu_y$ is the average of $y$, $\sigma_x^2$ and $\sigma_y^2$ are the variances of $x$ and $y$, $\sigma_{x y}$ is the covariance of $x$ and $y$, $c_1$ and $c_2$ are variables to stabilize the division with a weak denominator.

\begin{figure}[btph]
  \centering
  \begin{subfigure}{0.35\textwidth}
    \includegraphics[width=\textwidth]{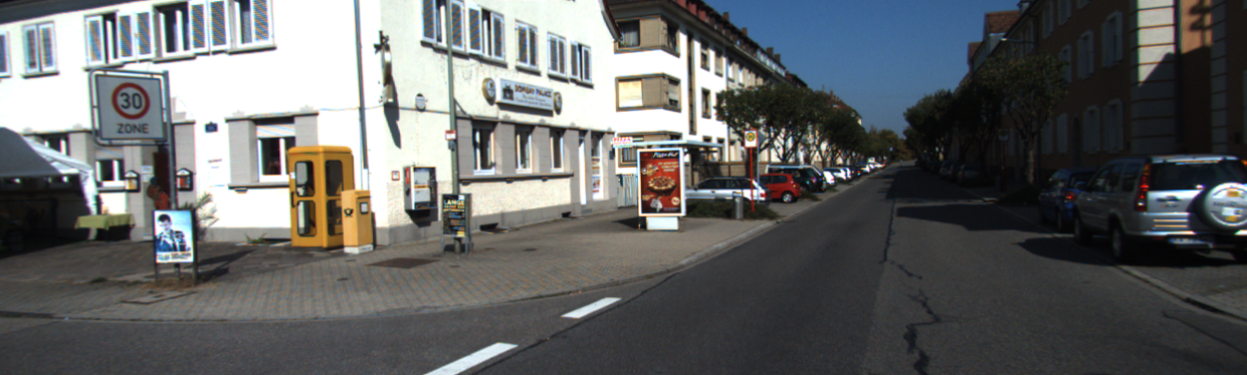} 
  \end{subfigure}
  \begin{subfigure}{0.35\textwidth}
    \includegraphics[width=\textwidth]{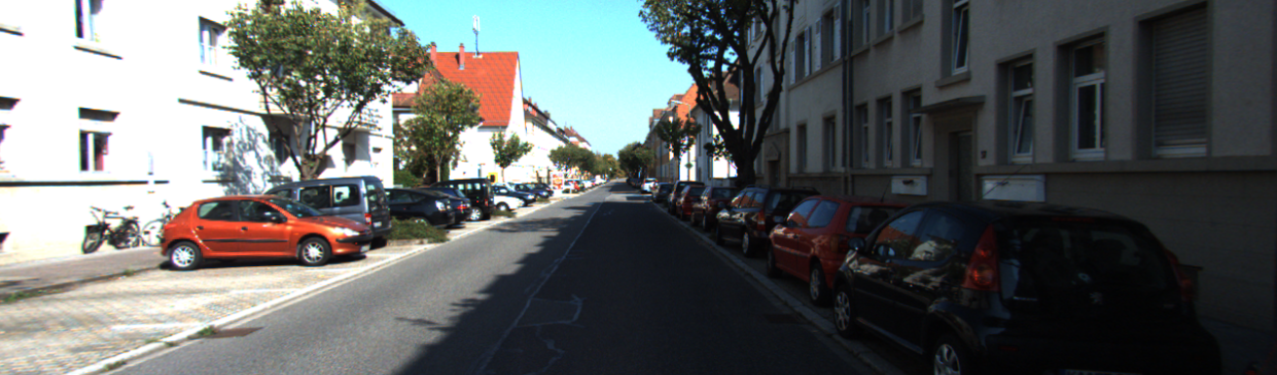} 
  \end{subfigure}
  \begin{subfigure}{0.35\textwidth}
    \includegraphics[width=\textwidth]{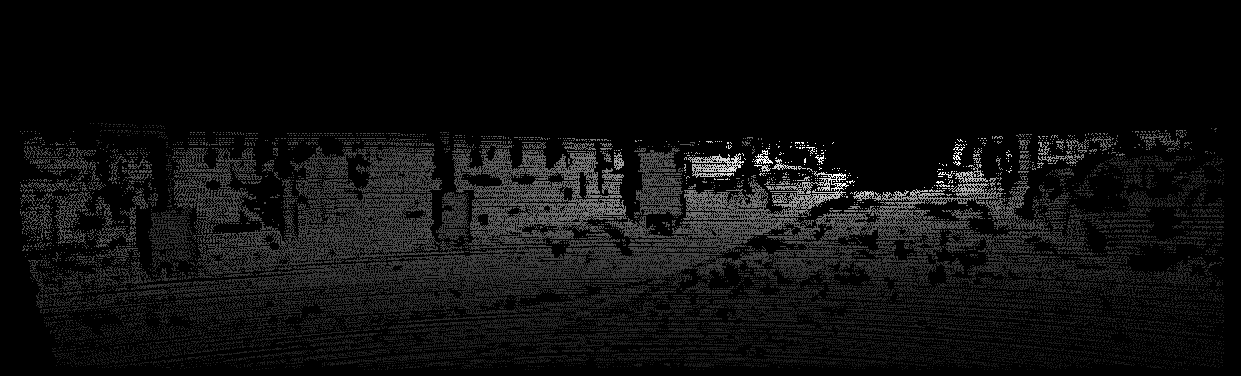}
    
  \end{subfigure}
  \begin{subfigure}{0.35\textwidth}
    \includegraphics[width=\textwidth]{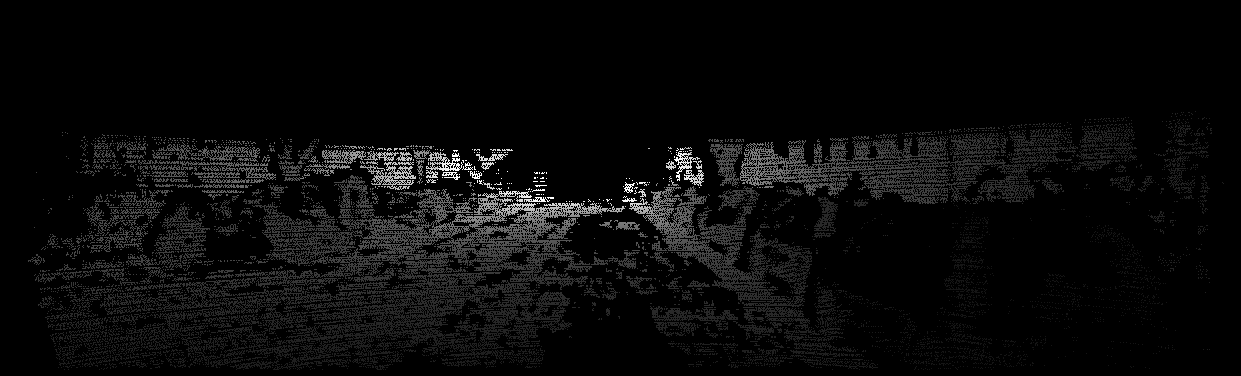}
    
  \end{subfigure}
  \begin{subfigure}{0.35\textwidth}
    \includegraphics[width=\textwidth]{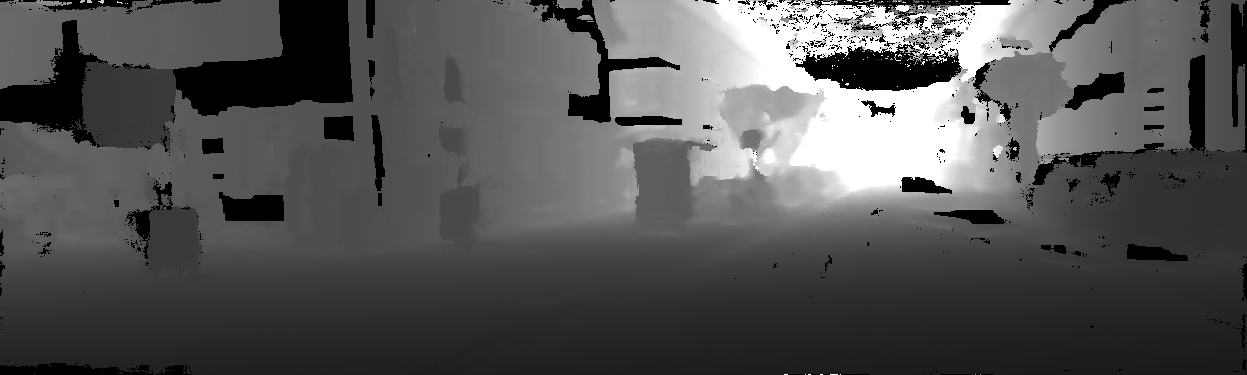}
  
  \end{subfigure}
    \begin{subfigure}{0.35\textwidth}
    \includegraphics[width=\textwidth]{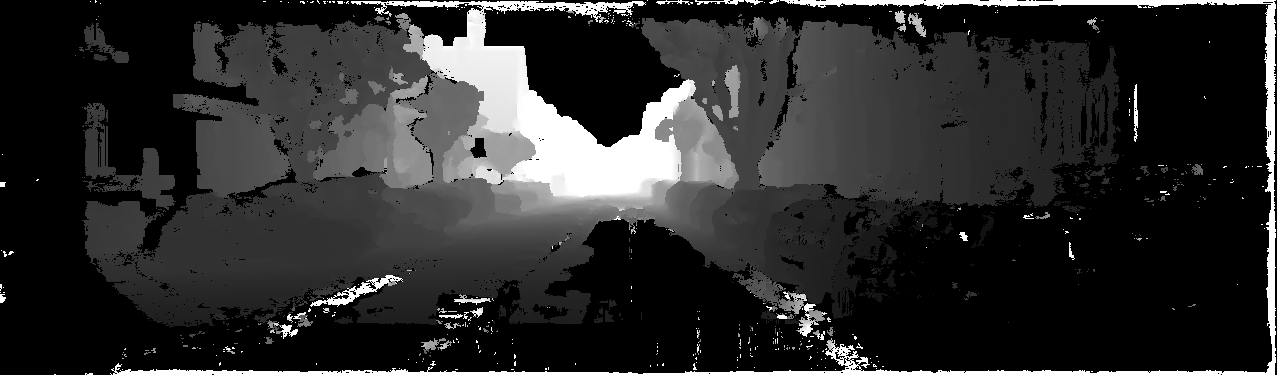}
    
  \end{subfigure}
  \begin{subfigure}{0.35\textwidth}
    \includegraphics[width=\textwidth]{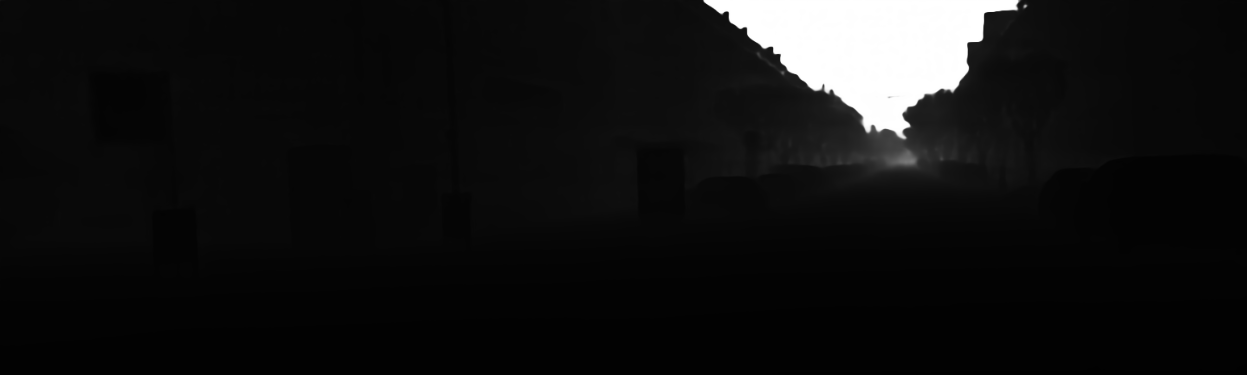}
    
  \end{subfigure}
  \begin{subfigure}{0.35\textwidth}
    \includegraphics[width=\textwidth]{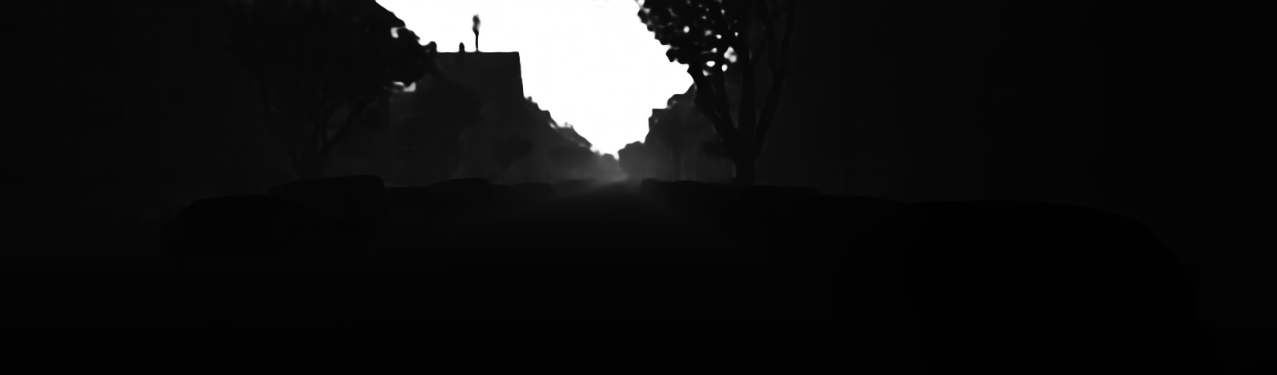}
    
  \end{subfigure}
  \begin{subfigure}{0.35\textwidth}
    \includegraphics[width=\textwidth]{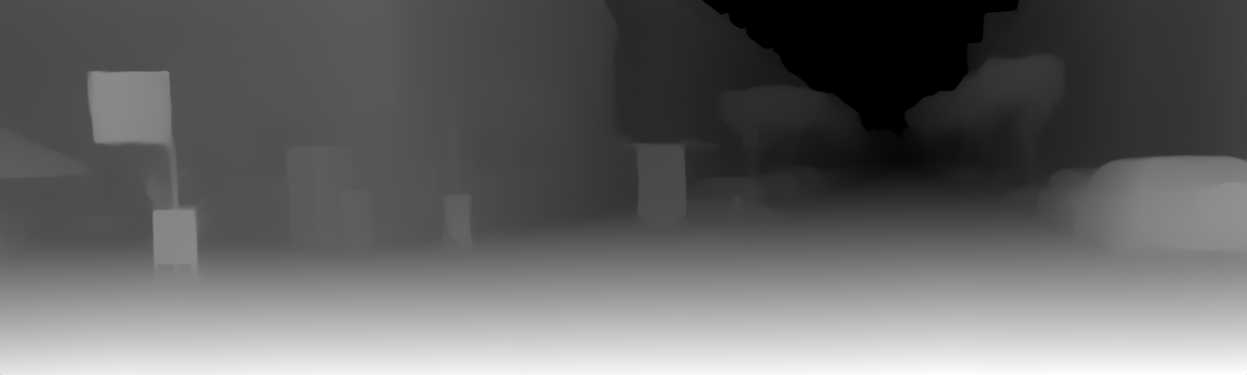}
    
  \end{subfigure}
  \begin{subfigure}{0.35\textwidth}
    \includegraphics[width=\textwidth]{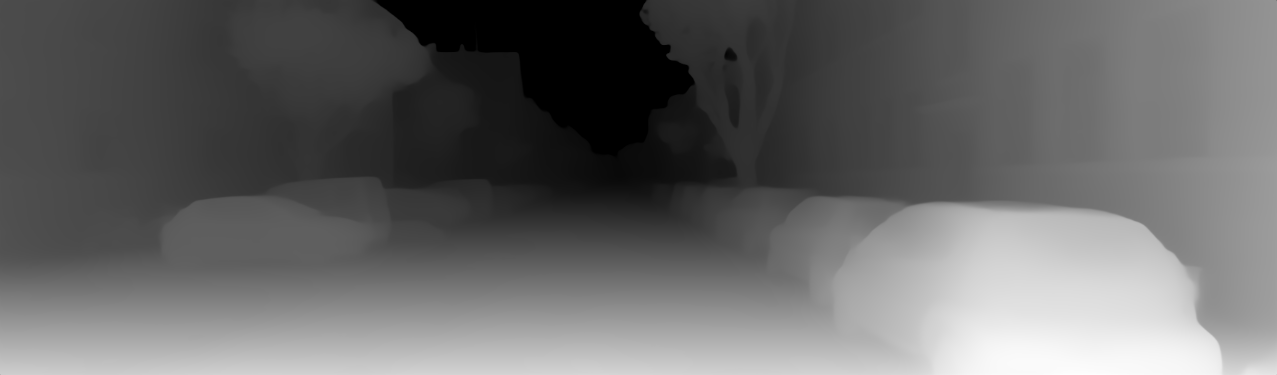} 
    
  \end{subfigure}
  \begin{subfigure}{0.35\textwidth}
    \includegraphics[width=\textwidth]{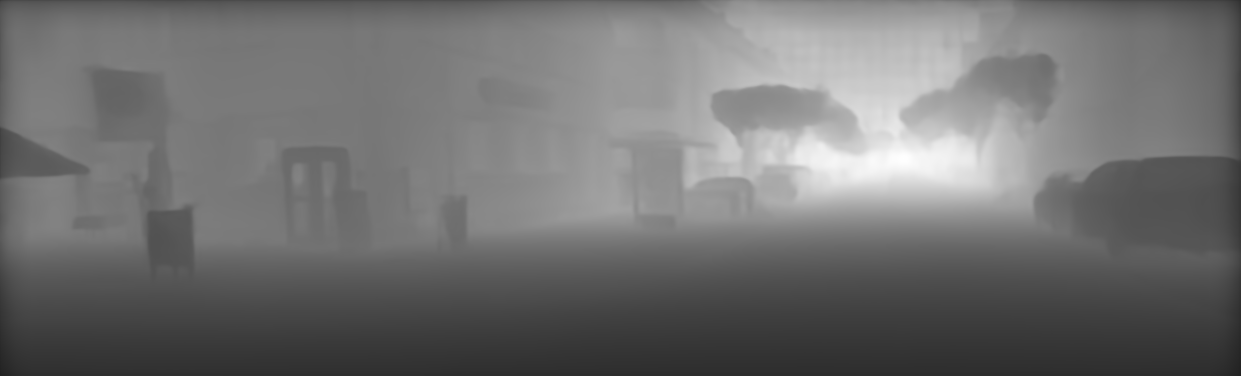} 
  \end{subfigure}
  \begin{subfigure}{0.35\textwidth}
    \includegraphics[width=\textwidth]{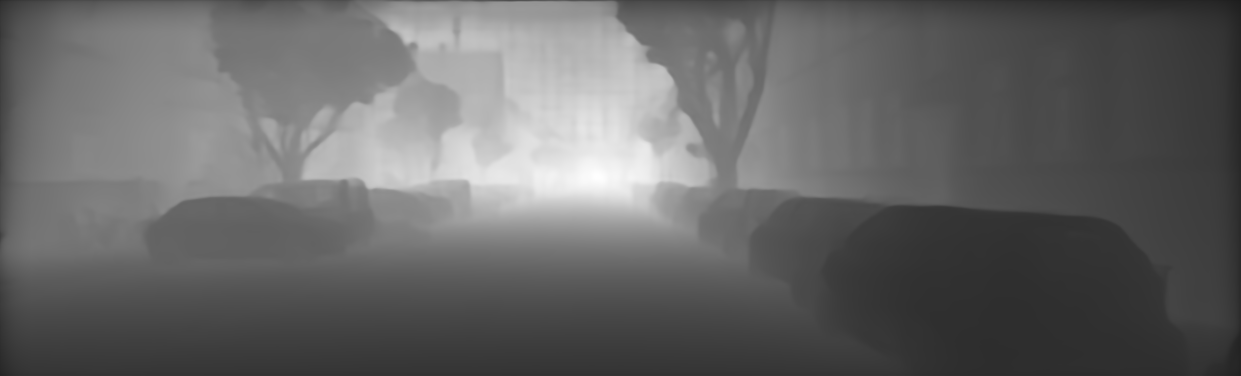}
  \end{subfigure}
  \caption{Depth Map Estimation Comparison. From top to bottom: the raw image, ground truth depth map, COLMAP dense depth map, EcoDepth, Depth Anything, and ZoeDepth.}
  \label{a}
\end{figure}
\subsection{Results and Discussions}
\subsubsection{Depth Evaluation on KITTI Datasets}
This set of experiments is meant to support our first assumption that the depth generated by COLMAP can be taken as depth ground truth in scenarios lacking ground truth information. This will further support our second assumption that we can take depth generated by COLMAP  as ground truth to evaluate depth obtained by monocular depth estimation models. All experiments were conducted using an NVIDIA A100 GPU.

We conducted rigorous evaluations of depth maps from COLMAP's dense depth reconstruction and various monocular depth estimation methods using the selected KITTI subsets with ground truth depth maps. The results, presented in Table \ref{tab2},
reveal that COLMAP consistently outperforms (or achieves on-the-par results) monocular methods across all metrics, with a $\delta_1$ value of 91\%, highlighting its closer approximation to ground truth. In scenarios where acquiring ground truth is cost-prohibitive, COLMAP's dense depth reconstruction proves to be a viable alternative. Additional details and visual comparisons of depth maps generated by different methods are illustrated in Figure \ref{a}. For instance, the COLMAP dense depth map, as shown in the second row of Figure \ref{a}, reveals a more precise delineation and representation of object edges compared to maps produced by monocular depth estimation methods.

\begin{table}[tbp]
\centering
\caption{Different Methods on Selected KITTI for Depth Estimation}
\label{tab2}
\begin{tabular}{|l|l|l|l|}
\hline
Methods & ${\delta_1 \uparrow}$   &RMSE $\downarrow$  & $\log _{10} \downarrow$ \\
\hline\hline
COLMAP&\textbf{0.910} &\textbf{0.017}&\textbf{0.042}\\
\hline
 ZoeDepth& {0.807} & 0.024&0.064\\
\hline
EcoDepth&{0.670} &0.017&0.080\\
\hline
Depth Anything&{0.184} &0.044&0.357\\
\hline
\end{tabular}
\end{table}

\subsubsection{Depth Evaluation on Removal NeRF Dataset}
\begin{figure}[tbhp]
  \centering

  \begin{subfigure}[b]{0.16\textwidth}
    \includegraphics[width=\textwidth]{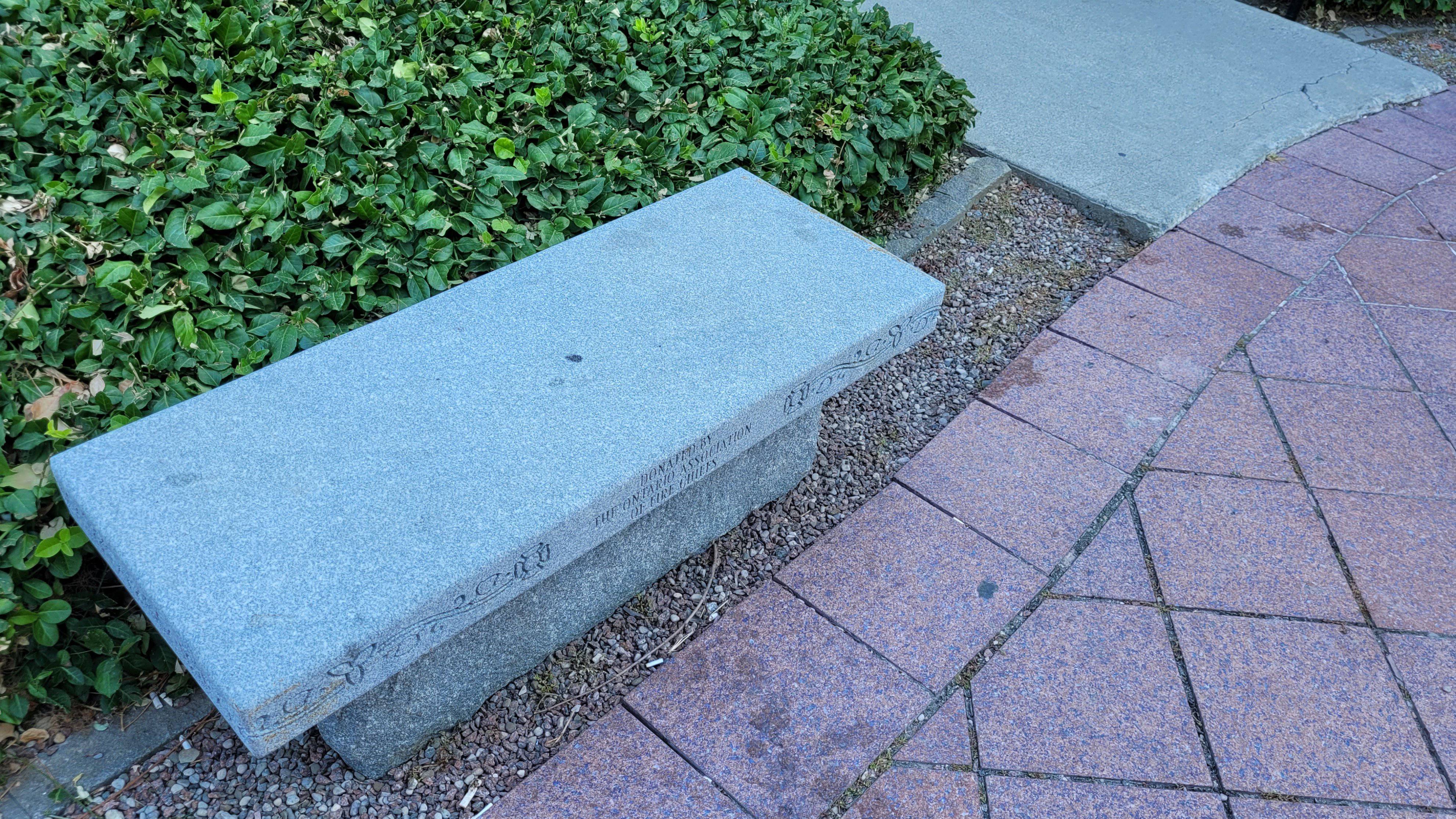}
    
  \end{subfigure}
  \begin{subfigure}[b]{0.16\textwidth}
    \includegraphics[width=\textwidth]{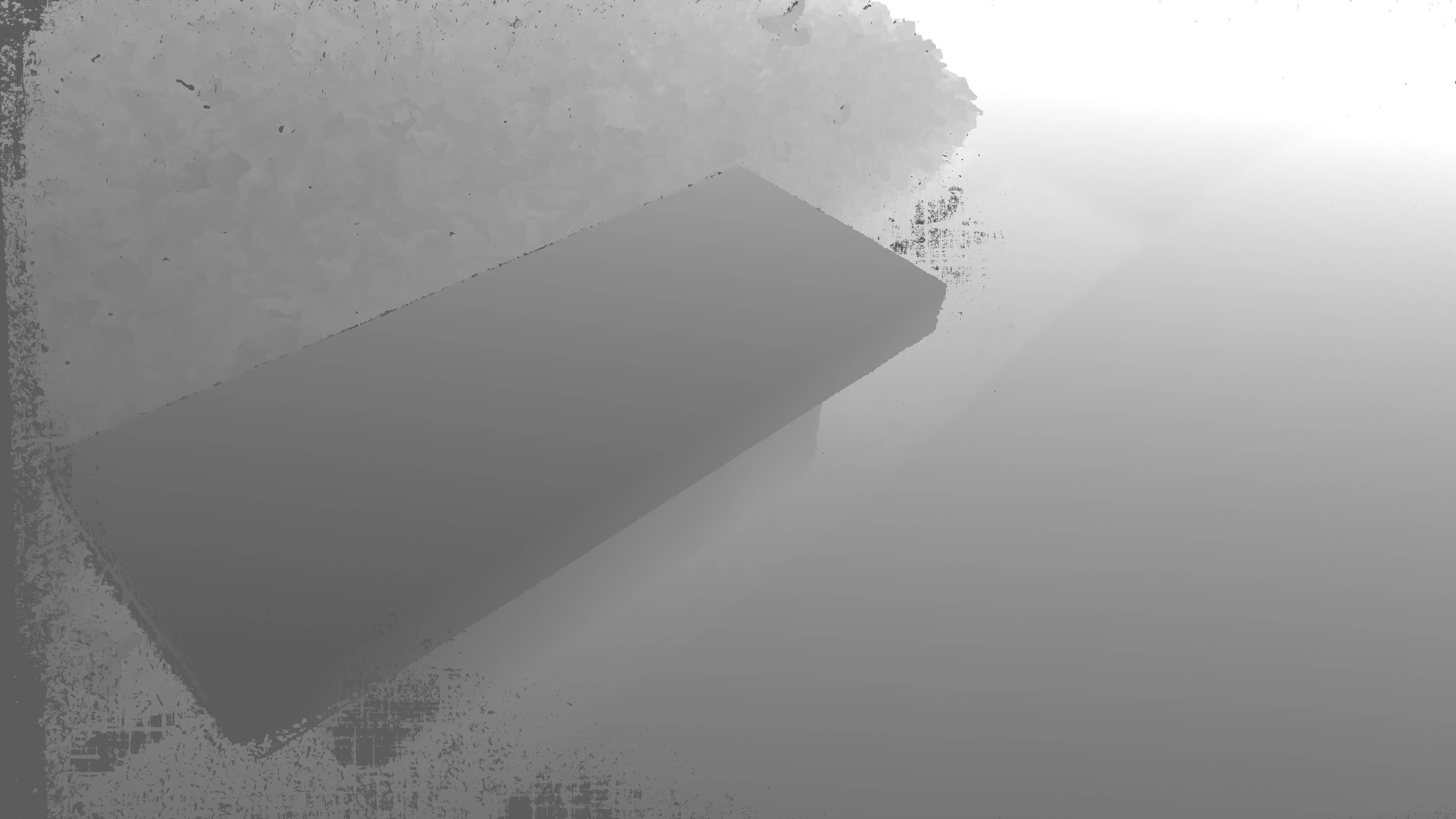}
    
  \end{subfigure}
  \begin{subfigure}[b]{0.16\textwidth}
    \includegraphics[width=\textwidth]{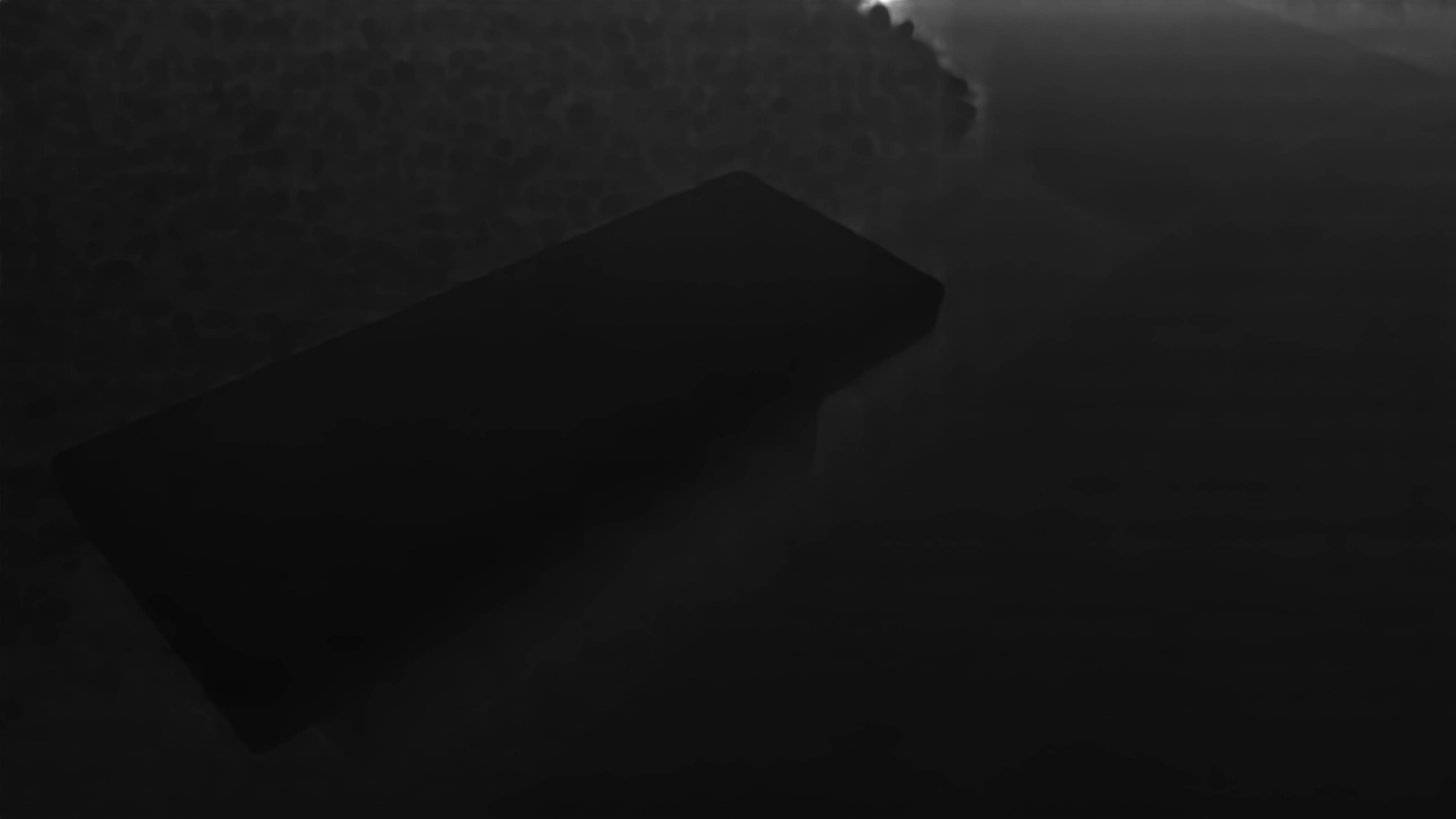}    
  \end{subfigure}
  \begin{subfigure}[b]{0.16\textwidth}
    \includegraphics[width=\textwidth]{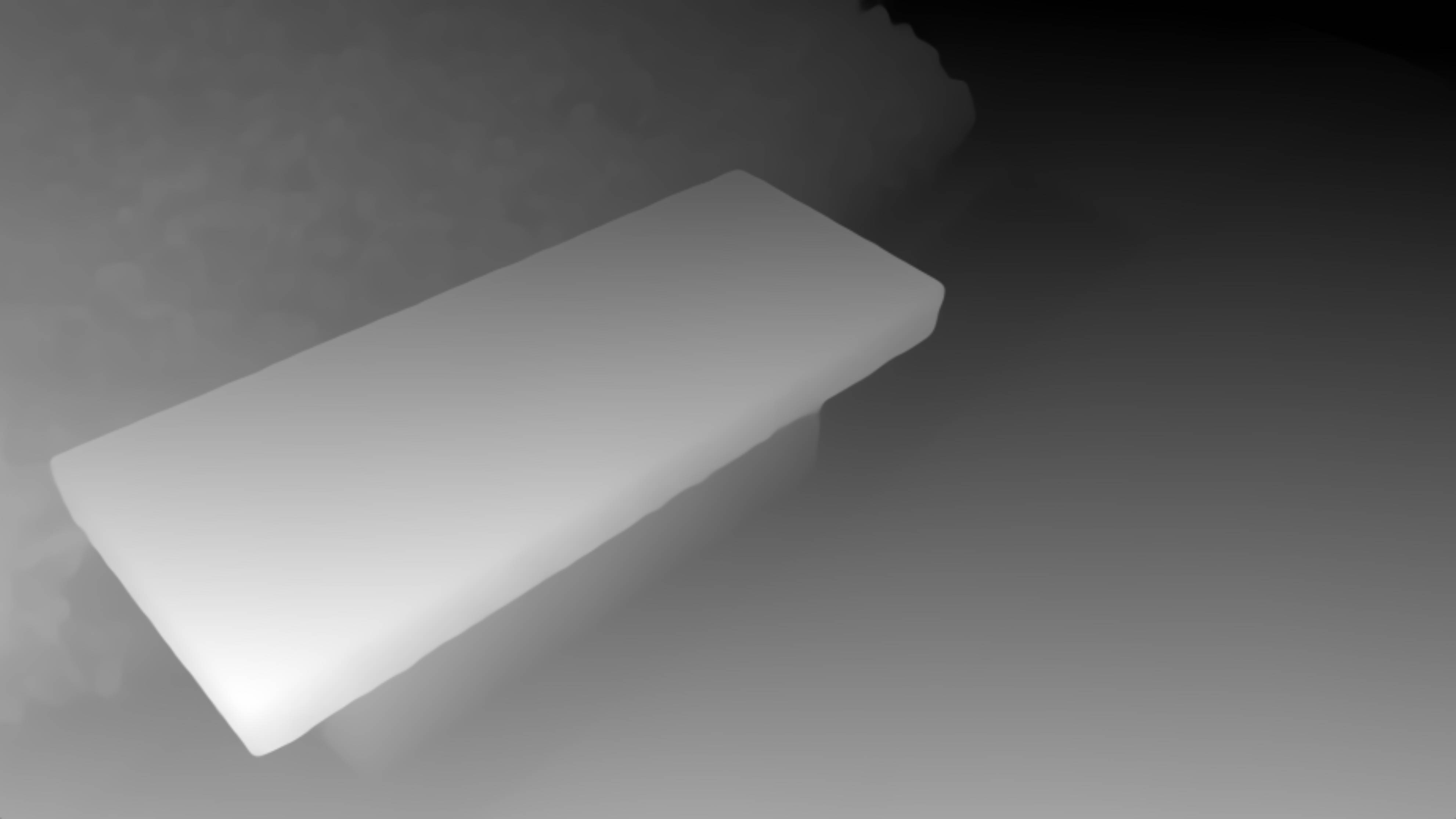}  
  \end{subfigure}
  \begin{subfigure}[b]{0.16\textwidth}
    \includegraphics[width=\textwidth]{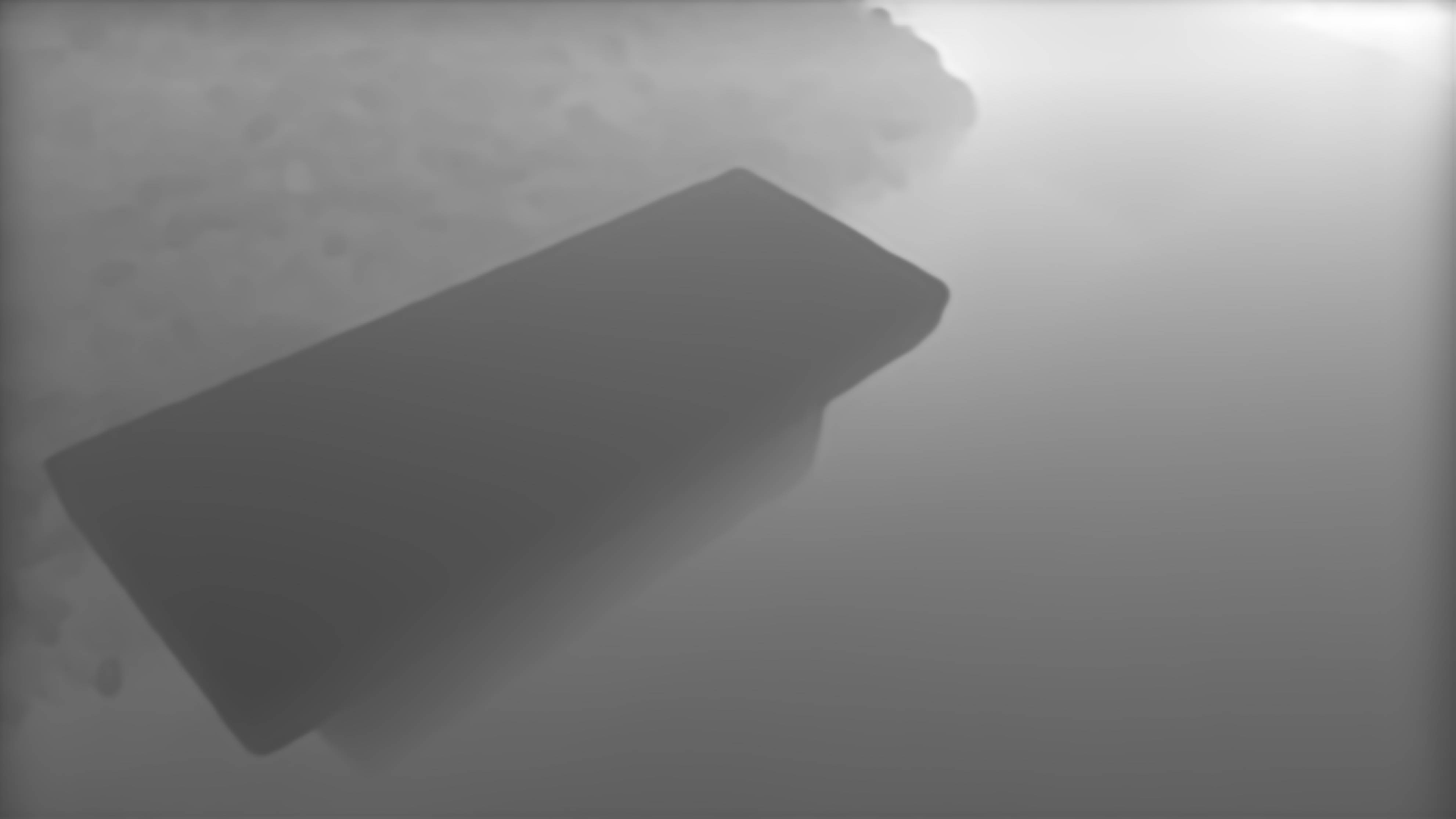}  
  \end{subfigure}
  \begin{subfigure}[b]{0.16\textwidth}
    \includegraphics[width=\textwidth]{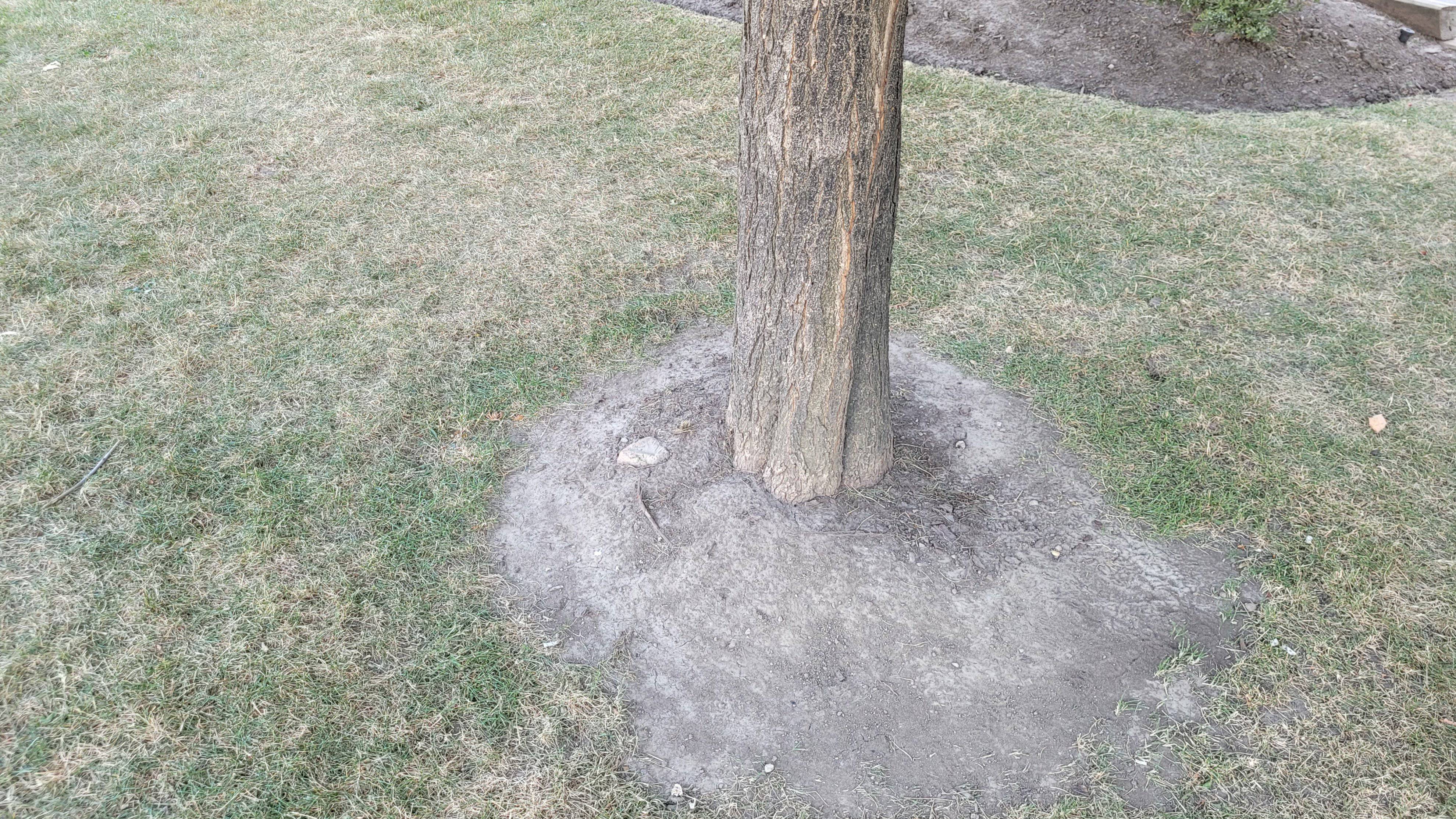}
    
  \end{subfigure}
  \begin{subfigure}[b]{0.16\textwidth}
    \includegraphics[width=\textwidth]{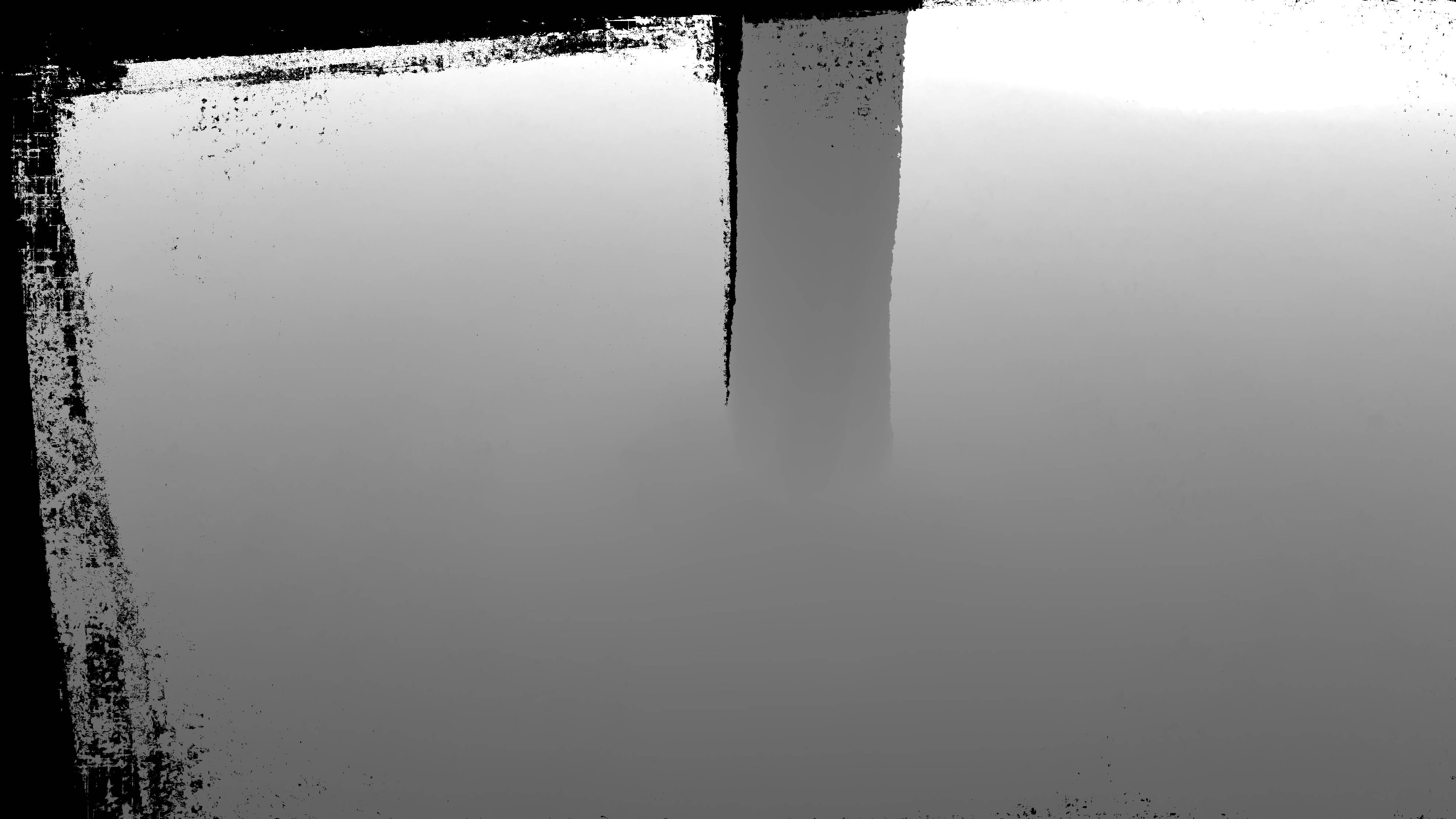}
    
  \end{subfigure}
  \begin{subfigure}[b]{0.16\textwidth}
    \includegraphics[width=\textwidth]{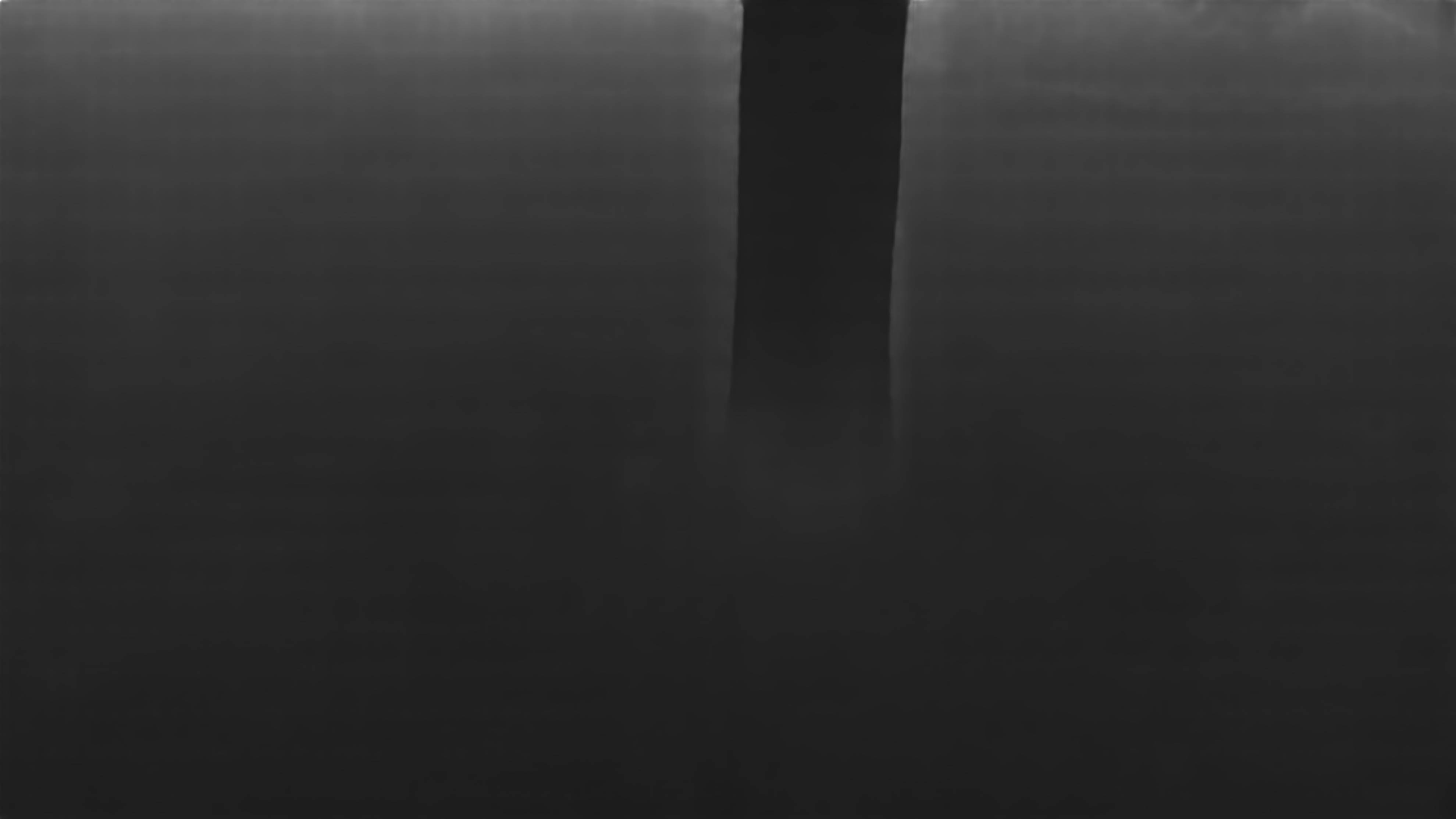}    
  \end{subfigure}
  \begin{subfigure}[b]{0.16\textwidth}
    \includegraphics[width=\textwidth]{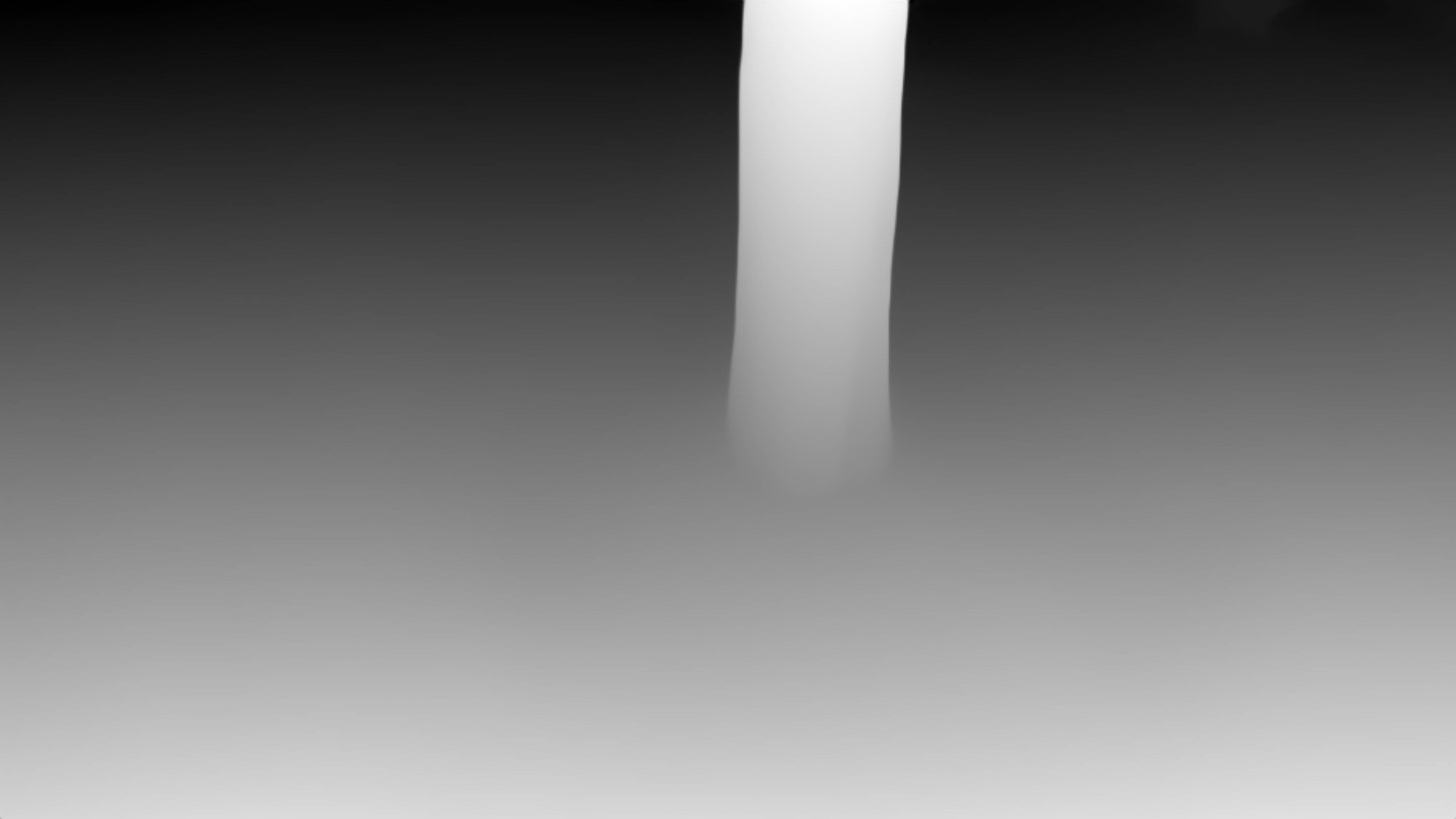}  
  \end{subfigure}
  \begin{subfigure}[b]{0.16\textwidth}
    \includegraphics[width=\textwidth]{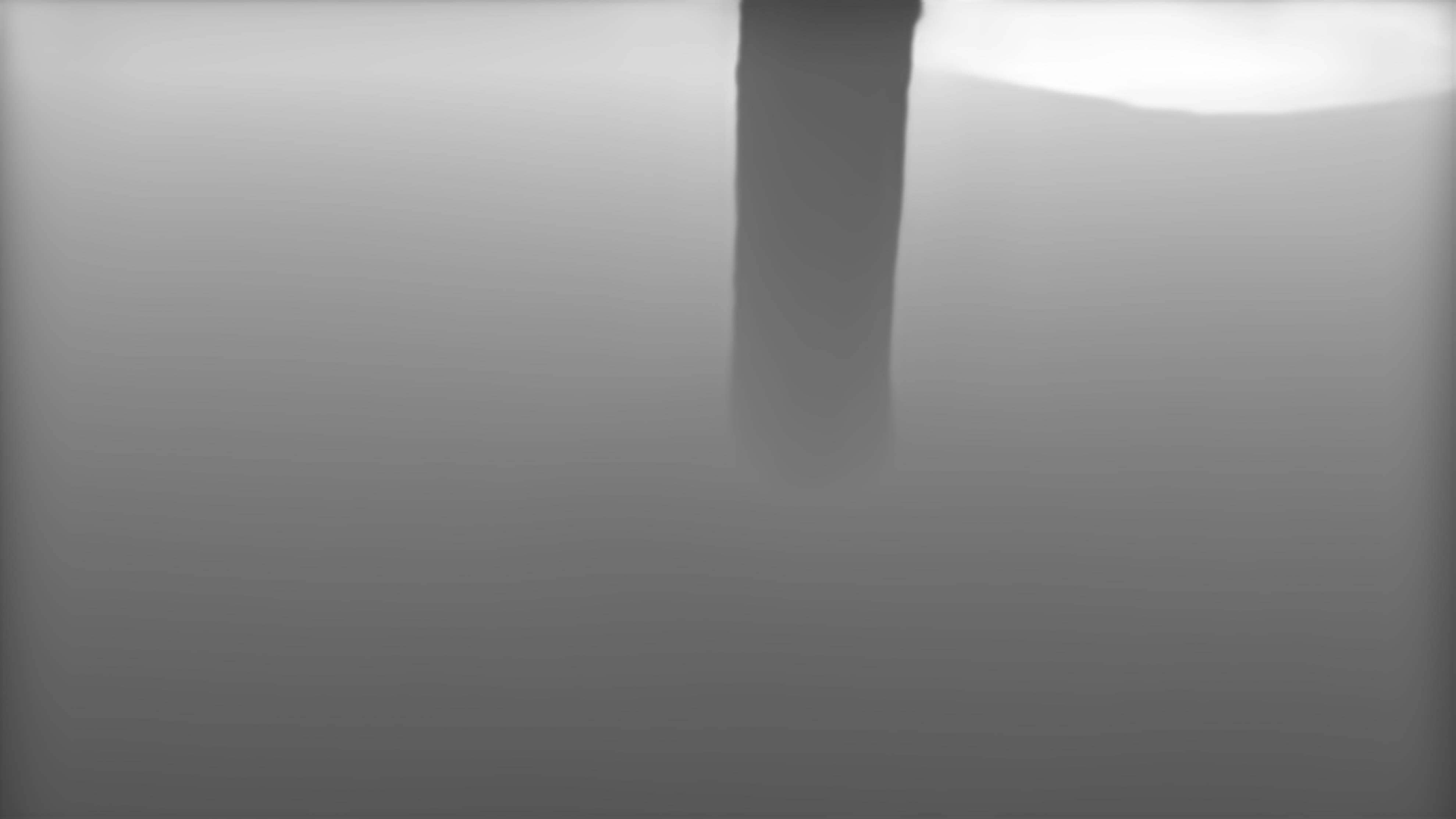}  
  \end{subfigure}
    \begin{subfigure}[b]{0.16\textwidth}
    \includegraphics[width=\textwidth]{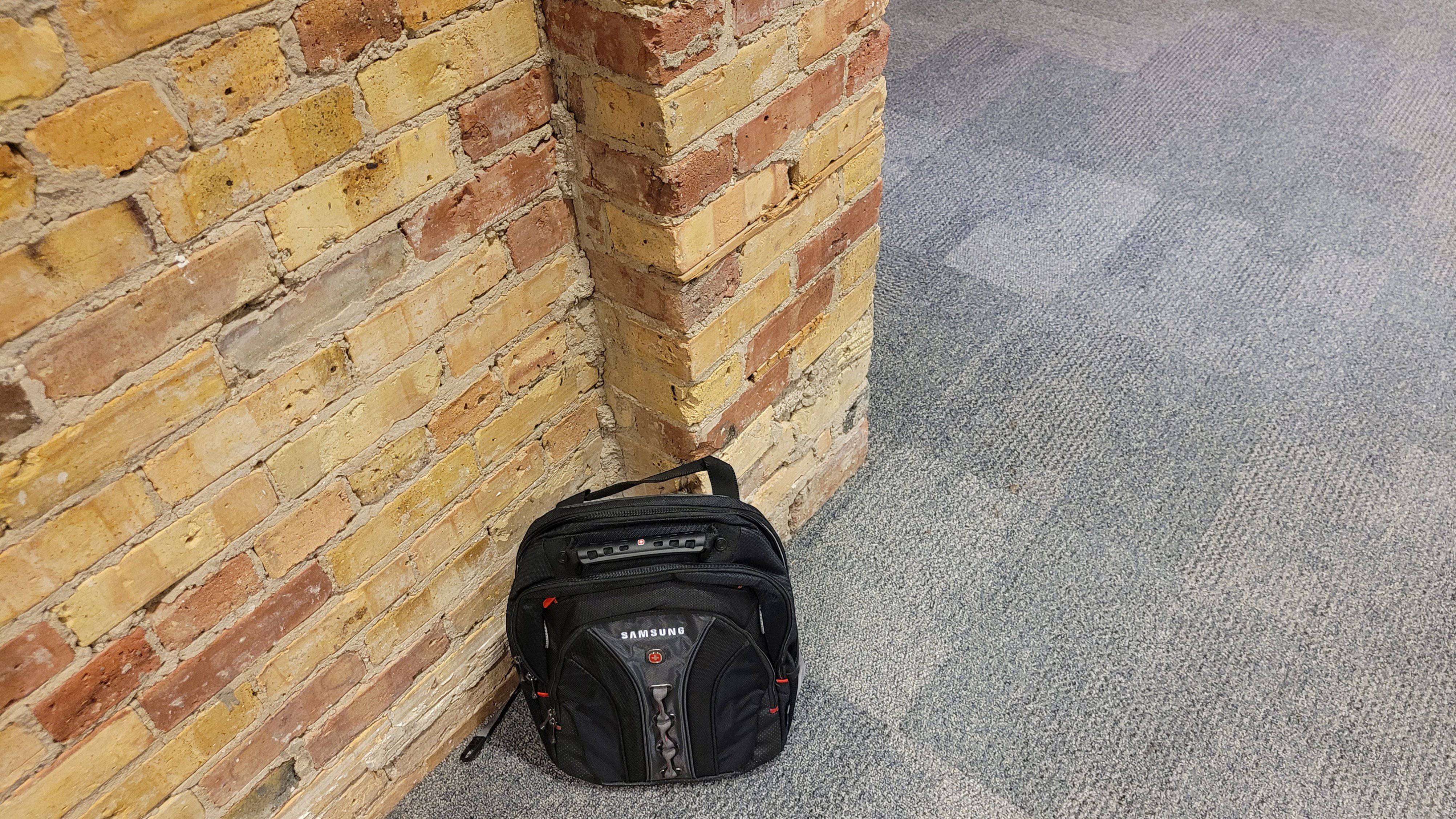}
    
  \end{subfigure}
  \begin{subfigure}[b]{0.16\textwidth}
    \includegraphics[width=\textwidth]{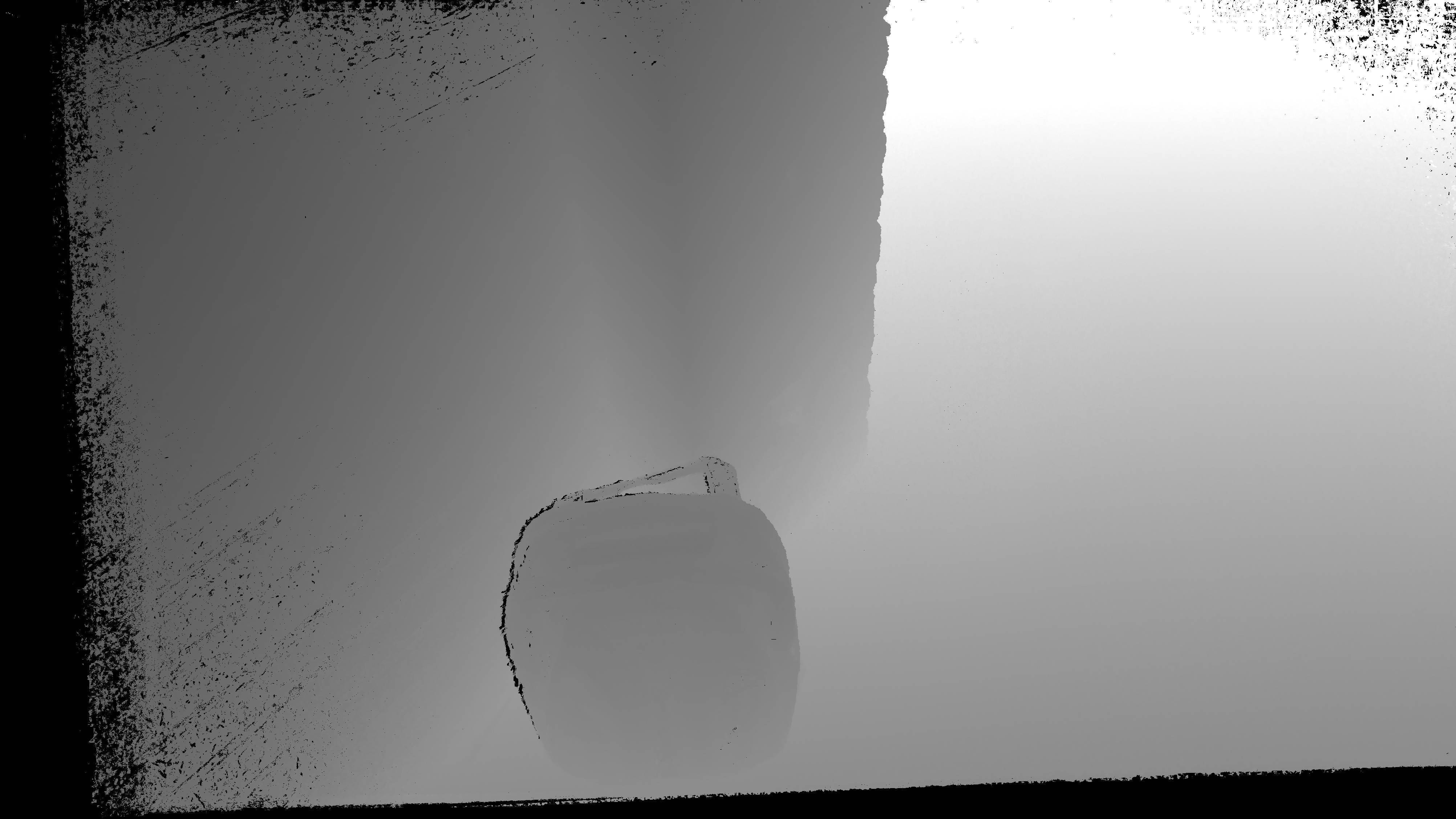}
    
  \end{subfigure}
  \begin{subfigure}[b]{0.16\textwidth}
    \includegraphics[width=\textwidth]{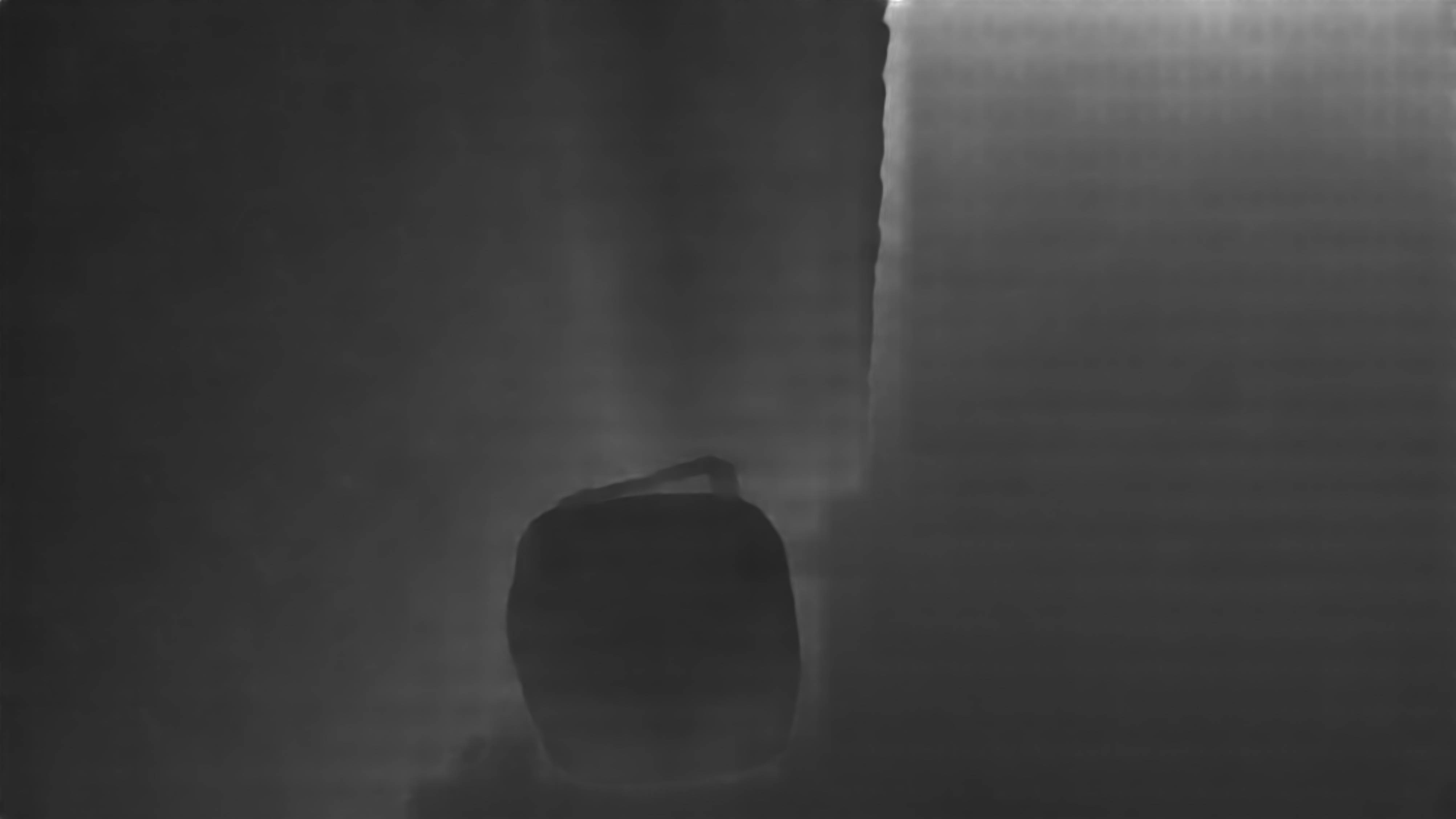}    
  \end{subfigure}
  \begin{subfigure}[b]{0.16\textwidth}
    \includegraphics[width=\textwidth]{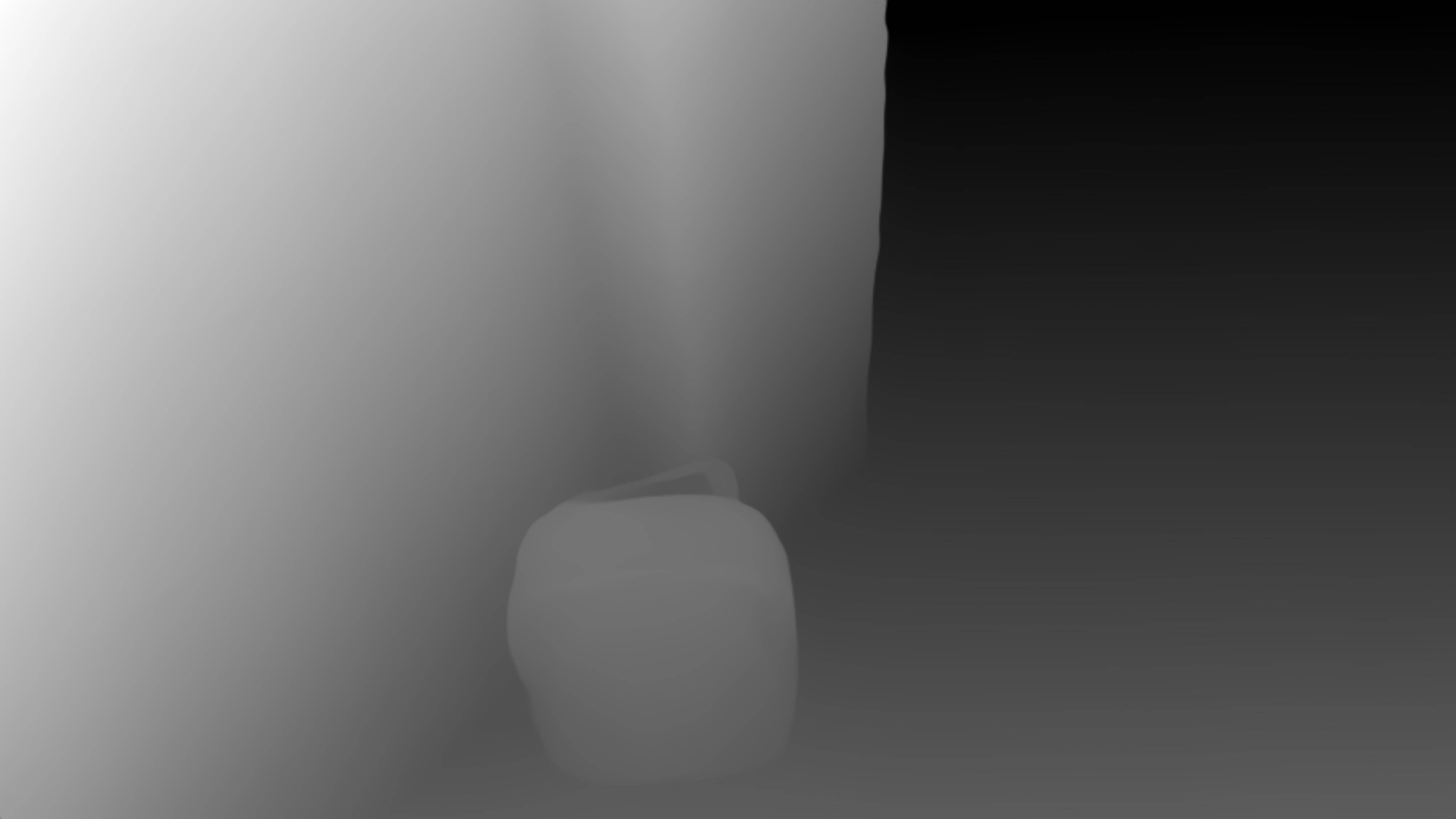}  
  \end{subfigure}
  \begin{subfigure}[b]{0.16\textwidth}
    \includegraphics[width=\textwidth]{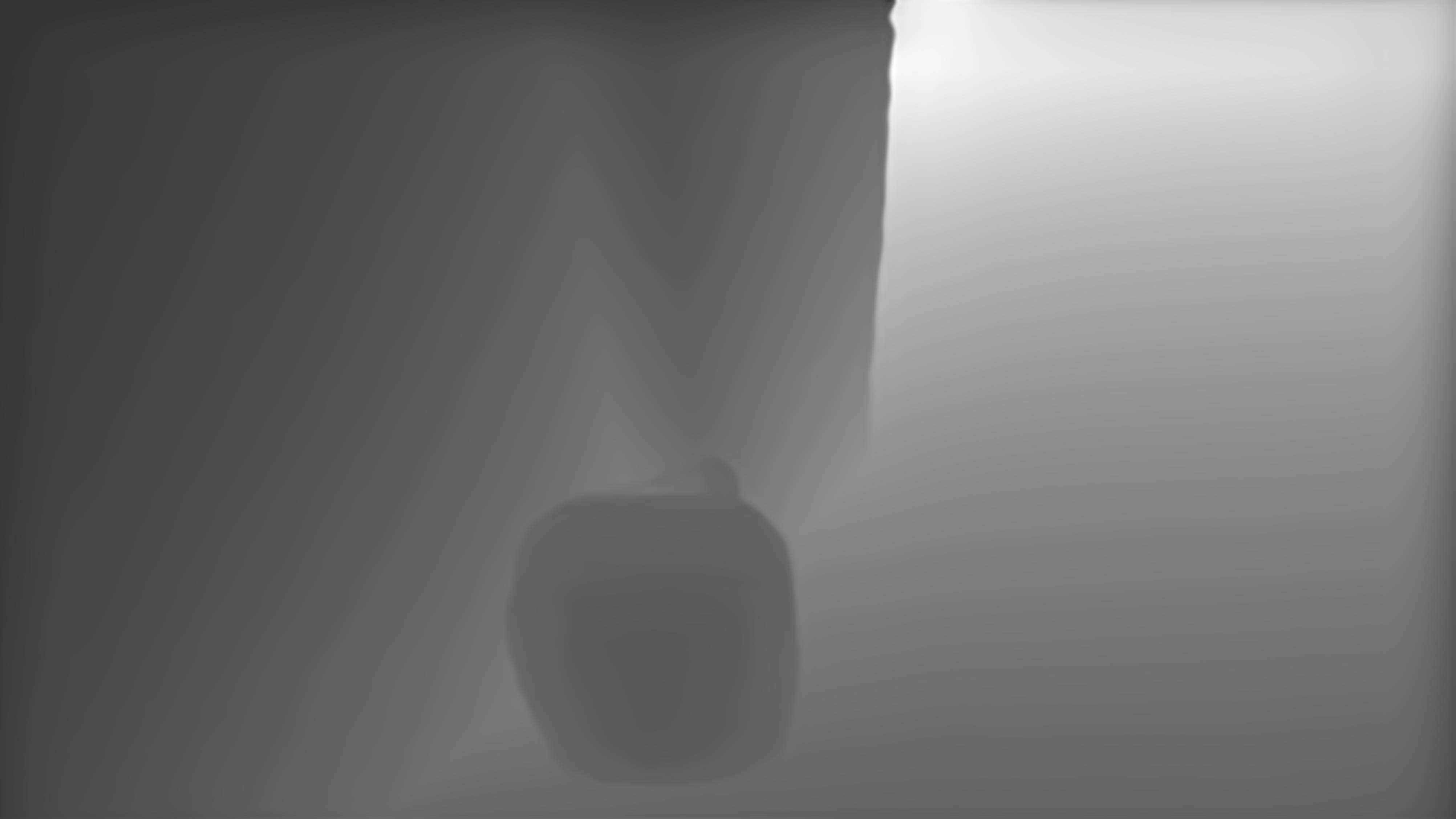}  
  \end{subfigure}
  \begin{subfigure}[b]{0.16\textwidth}
    \includegraphics[width=\textwidth]{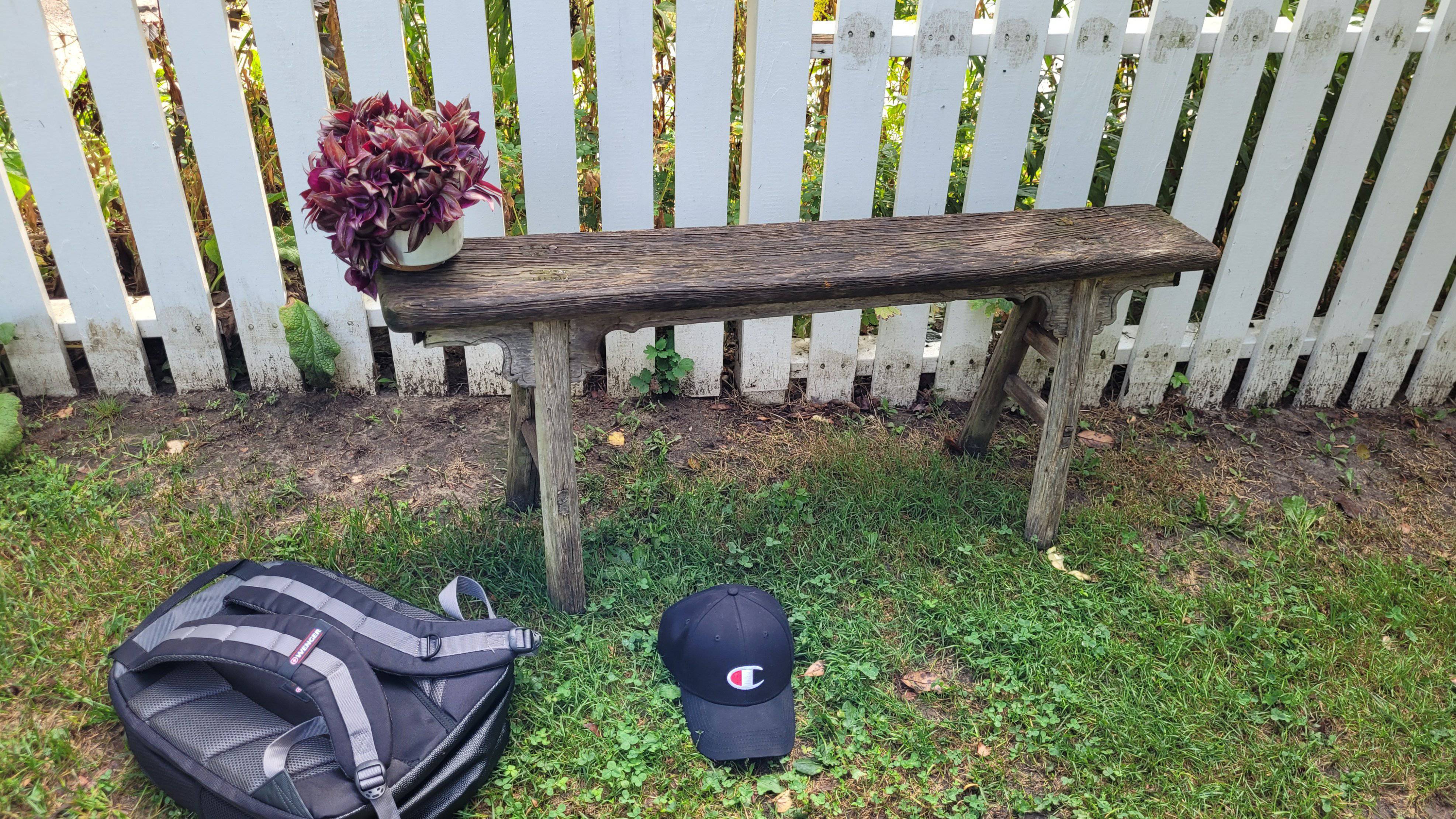}
    
  \end{subfigure}
  \begin{subfigure}[b]{0.16\textwidth}
    \includegraphics[width=\textwidth]{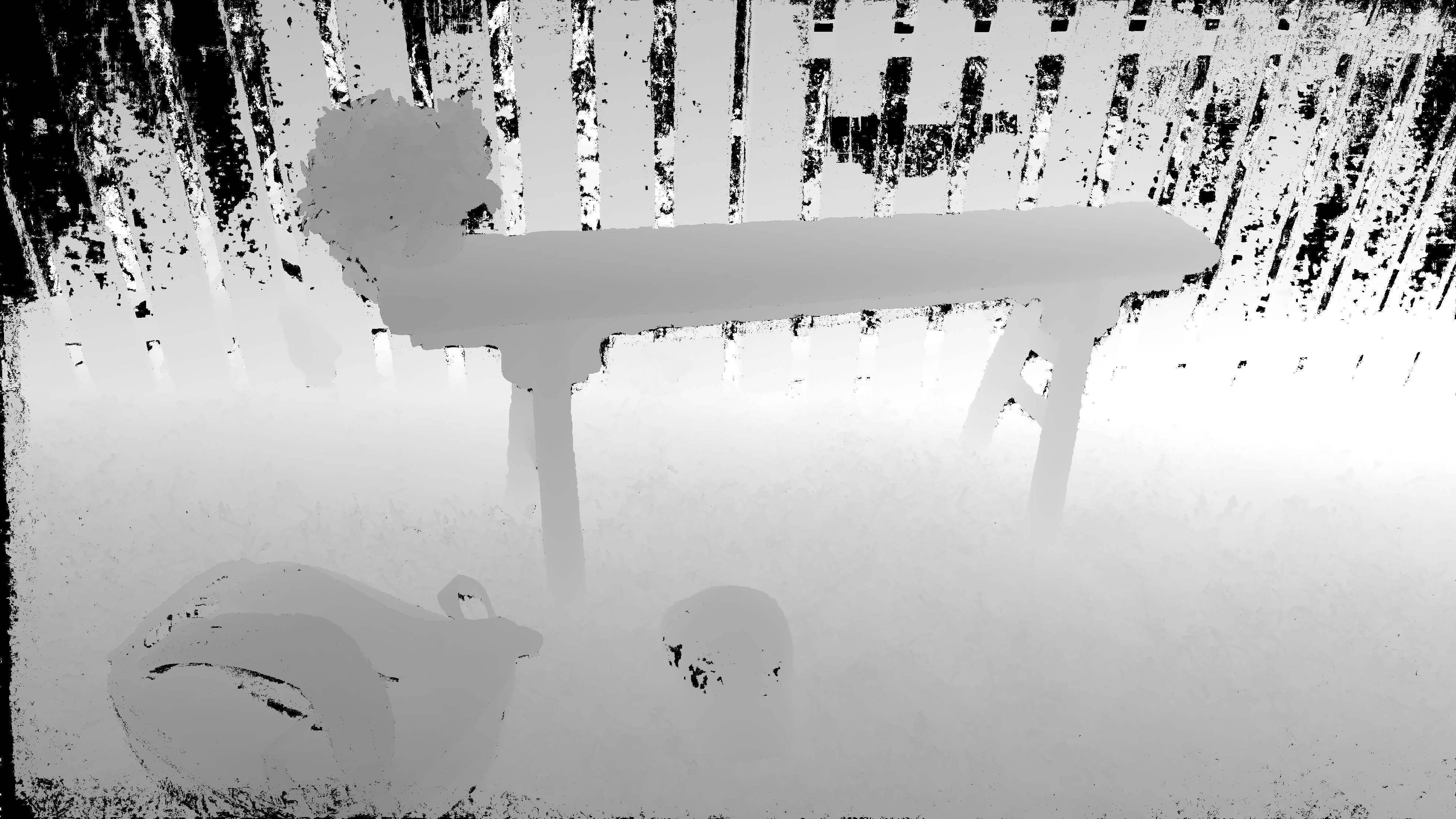}
    
  \end{subfigure}
  \begin{subfigure}[b]{0.16\textwidth}
    \includegraphics[width=\textwidth]{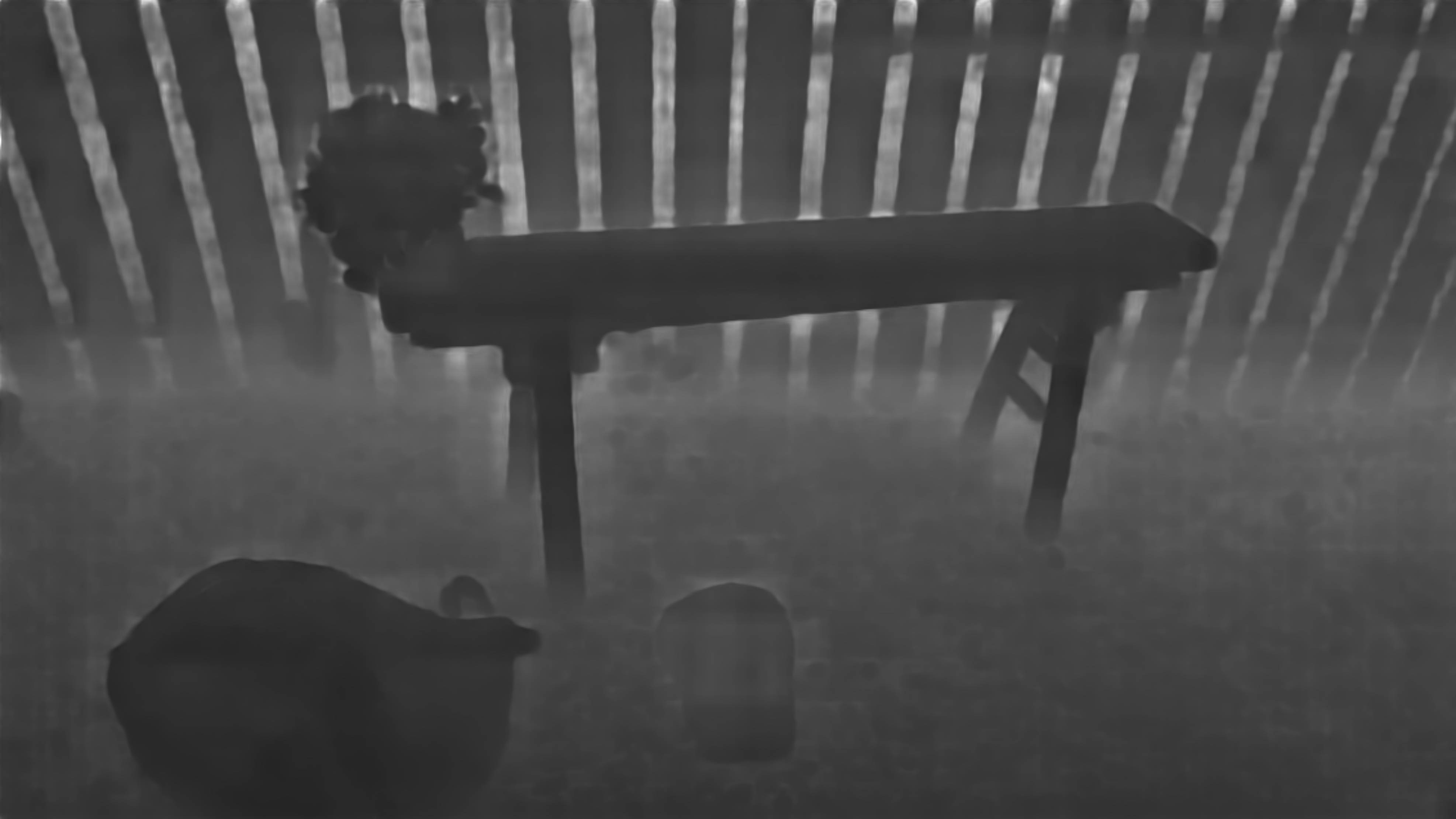}  
  \end{subfigure}
  \begin{subfigure}[b]{0.16\textwidth}
    \includegraphics[width=\textwidth]{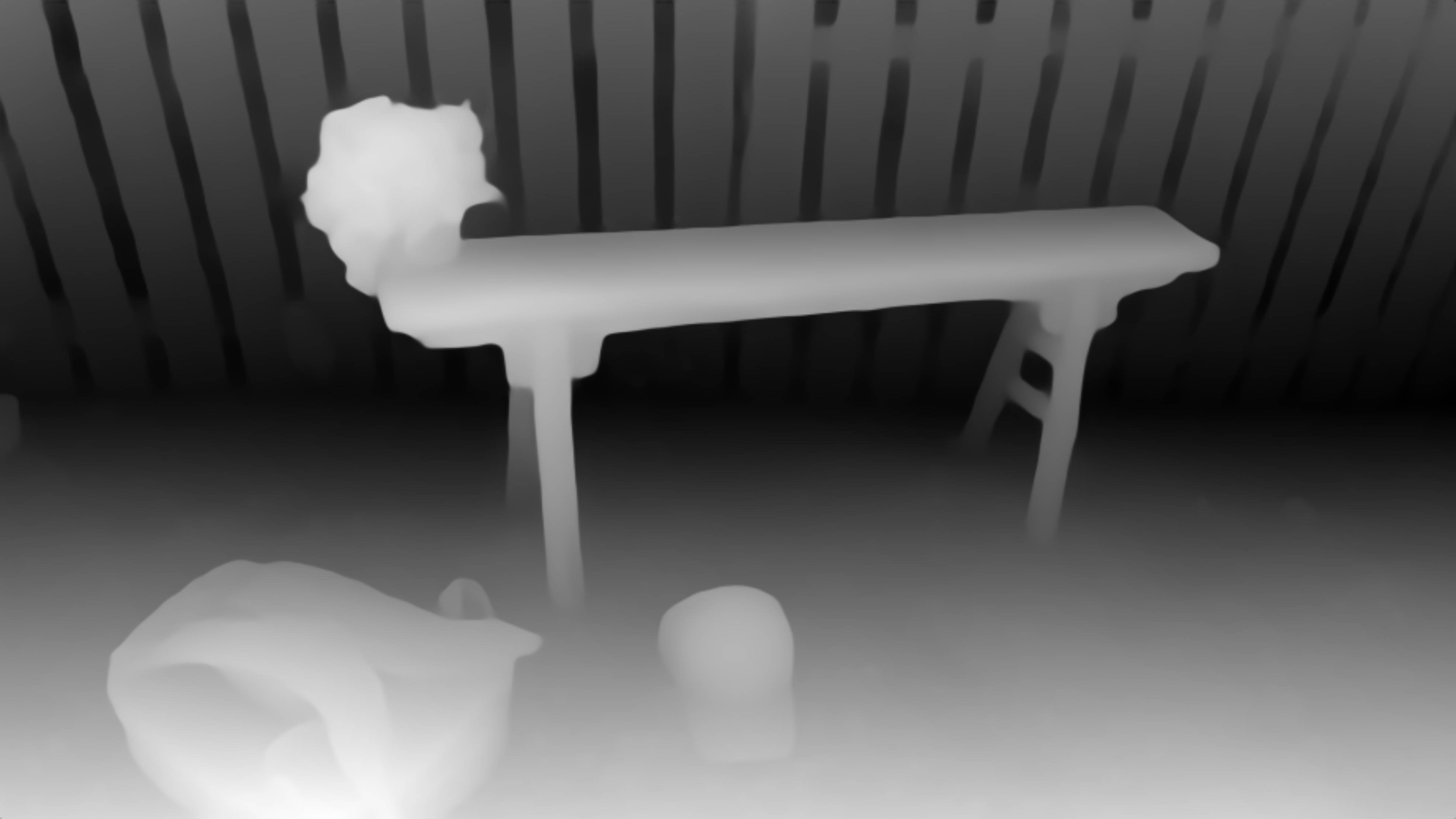}
  \end{subfigure}
  \begin{subfigure}[b]{0.16\textwidth}
    \includegraphics[width=\textwidth]{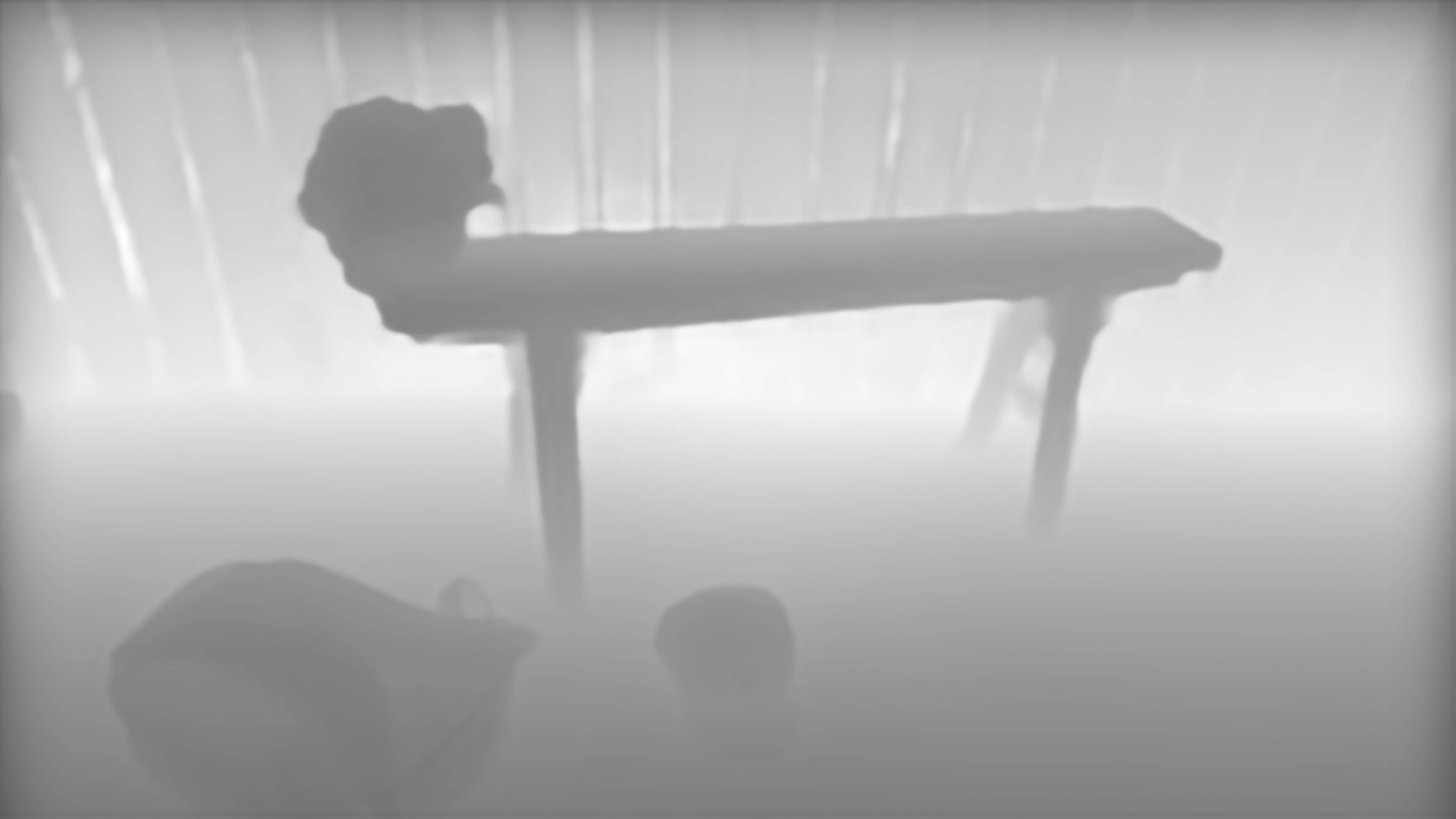}
  \end{subfigure}
  \caption{Estimated Depth Map Comparison. From left to right columns: input images; depth maps from COLMAP; depth maps from EcoDepth; depth maps from Depth Anything; depth maps from ZoeDepth.}
  \label{b}
\end{figure}

Leveraging findings from COLMAP depth studies, we generated ground truth depth maps for the Removal NeRF Dataset from SpinNeRF. We conducted a comprehensive evaluation using several monocular depth estimation methods, with detailed metrics presented in Table \ref{tab3}. Density in Table \ref{tab3} is defined as the proportion of pixels that contain non-zero values, which serves as a quantitative measure of the coverage and completeness of depth across the scene. Due to the significant variation in depth distances within our dataset, we mainly use the $\delta_1$ metric to evaluate the adaptability and robustness of methods across different depth scales. Our results highlight ZoeDepth as the best method for depth map acquisition in the Removal NeRF dataset, achieving over 94\% overlap with true depth values, significantly outperforming EcoDepth and DepthAnything. The depth maps results in Figure \ref{b} further illustrate ZoeDepth's enhanced detail capture, notably in object shape and edge delineation.

\begin{table}[h]
\centering
\caption{Monocular Depth Estimation.`D' stands for `Dataset'.}\label{tab3}
\begin{tabular}{|l|l|l|l|l|l|l|l|l|l|l|l|}
\hline
Models & Metric & D-1 & D-2 & D-3 & D-4 & D-5 & D-6 & D-7 & D-8 & D-9 & D-10 \\
\hline\hline
\multirow{4}{*}{ZoeDepth} 
& ${\delta_1 \uparrow}$ & \textbf{0.53} & \textbf{0.99} & \textbf{1.00} & \textbf{0.95} & \textbf{0.98} & \textbf{1.00} & \textbf{0.99} & \textbf{0.99} & \textbf{0.99} & \textbf{0.99} \\
& RMSE $\downarrow$ & 0.10 & 0.03 & 0.03 & 0.07 & 0.04 & 0.05 & 0.03 & 0.06 & 0.04 & 0.06 \\
& $\log _{10} \downarrow$ & 0.15 & 0.03 & 0.02 & 0.29 & 0.02 & 0.02 & 0.02 & 0.03 & 0.02 & 0.02 \\
& Density& 0.97 & 0.99 & 0.99 & 0.89 & 0.98 & 0.99 & 0.98 & 0.85 & 0.91 & 0.81 \\
\hline
\multirow{4}{*}{Depth Anything} 
& $\delta_1 \uparrow$ & 0.06 & 0.26 & 0.23 & 0.33 & 0.36 & 0.27 & 0.25 & 0.15 & 0.22 & 0.26 \\
& RMSE $\downarrow$ & 0.17 & 0.31 & 0.29 & 0.26 & 0.45 & 0.33 & 0.33 & 0.64 & 0.36 & 0.50 \\
& $\log _{10} \downarrow$ & 0.64 & 0.21 & 0.21 & 0.20 & 0.19 & 0.16 & 0.25 & 0.34 & 0.22 & 0.22 \\
& Density & 0.97 & 0.99 & 0.99 & 0.89 & 0.98 & 0.99 & 0.98 & 0.85 & 0.91 & 0.81 \\
\hline
\multirow{4}{*}{ECoDepth} 
& $\delta_1 \uparrow$ & 0.50 & 0.95 & 0.99 & 0.82 & 0.96 & 0.92 & 0.62 & 0.89 & 0.94 & 0.63 \\
& RMSE $\downarrow$ & 0.08 & 0.05 & 0.03 & 0.09 & 0.07 & 0.07 & 0.11 & 0.12 & 0.06 & 0.17 \\
& $\log _{10} \downarrow$ & 0.13 & 0.05 & 0.02 & 0.06 & 0.04 & 0.03 & 0.08 & 0.05 & 0.04 & 0.08 \\
& Density & 0.97 & 0.99 & 0.99 & 0.89 & 0.98 & 0.99 & 0.98 & 0.85 & 0.91 & 0.81 \\
\hline
\end{tabular}
\end{table}

\subsubsection{Different Depth Priors Comparison in Removal NeRF}
\begin{figure}[tbhp]
  \centering
  
  \begin{subfigure}[b]{0.2\textwidth}
    \includegraphics[width=\textwidth]{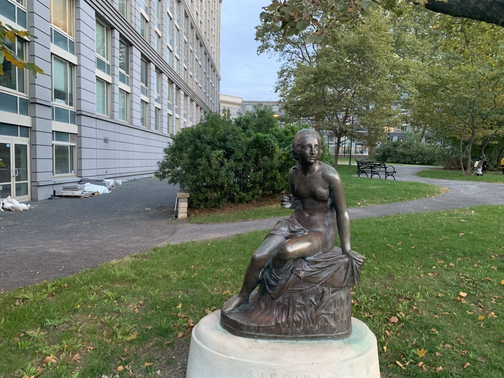}
  \end{subfigure}
  \begin{subfigure}[b]{0.2\textwidth}
    \includegraphics[width=\textwidth]{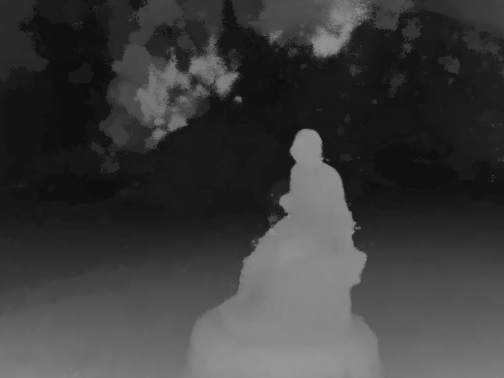}
  \end{subfigure}
    \begin{subfigure}[b]{0.2\textwidth}
    \includegraphics[width=\textwidth]{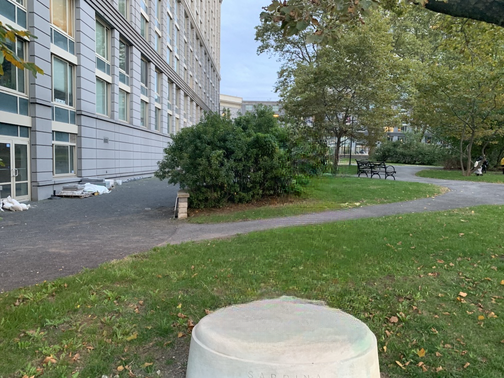}
  \end{subfigure}
  \begin{subfigure}[b]{0.2\textwidth}
    \includegraphics[width=\textwidth]{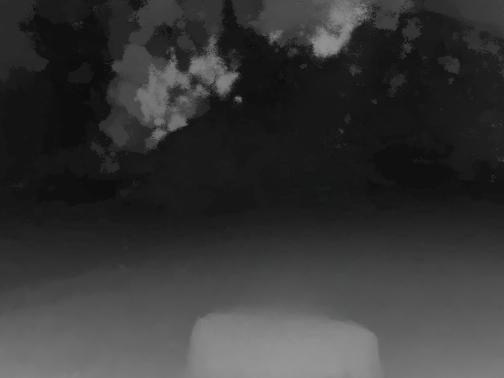}
  \end{subfigure}
    \begin{subfigure}[b]{0.2\textwidth}
    \includegraphics[width=\textwidth]{figures/spinnerf/raw_image.png}
  \end{subfigure}
  \begin{subfigure}[b]{0.2\textwidth}
    \includegraphics[width=\textwidth]{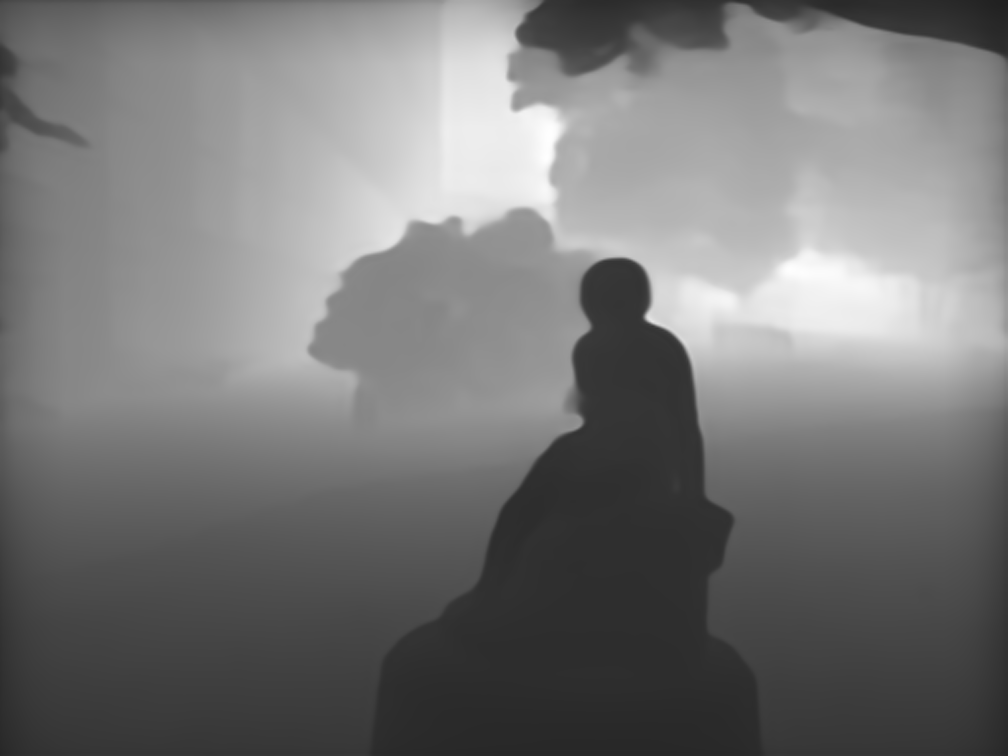}
  \end{subfigure}
    \begin{subfigure}[b]{0.2\textwidth}
    \includegraphics[width=\textwidth]{figures/spinnerf/lama_raw_image.png}
  \end{subfigure}
  \begin{subfigure}[b]{0.2\textwidth}
    \includegraphics[width=\textwidth]{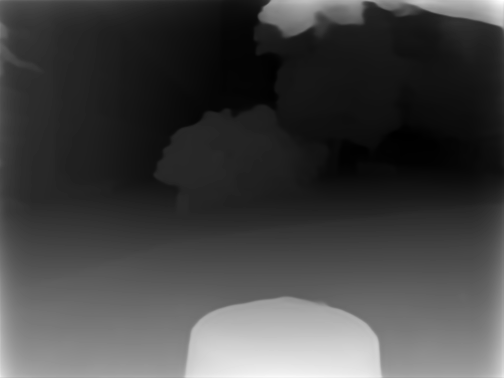}
  \end{subfigure}
  \caption{Depth Map Comparison on Input Image and Inpainted Image. \textit{Top Row}, from left to right: the input image; the depth map obtained by DSNeRF; the inpainted image; the inpainted depth map. \textit{Bottom Row}, from left to right: the input image; the depth map obtained by ZoeDepth; the inpainted image; the inpainted depth map}.
  \label{Depth Map Comparison on Input Image and Inpainted Image}
\end{figure}

To further assess the impact of various depth estimation models on object removal within NeRF, we continued our experiments using SpinNeRF as a benchmark, rigorously evaluating its performance with both its original pipeline where DSNeRF was used for complete depth acquisition and when augmented with ZoeDepth monocular depth estimation. Detailed comparison results are presented in Table \ref{tab4}. We found that using ZoeDepth instead of the complete depth provided by DSNeRF not only improved the PSNR by $6.87\%$ but also significantly reduced the depth estimation time consumption from 44.5 seconds per frame to 0.58 seconds per frame, making our pipeline appealing to real-time applications such as human-robot collaboration. Additionally, comparisons of inpainting results using different depth priors are shown in Figure \ref{Depth Map Comparison on Input Image and Inpainted Image}. Our findings indicate that ZoeDepth excels in capturing the shape, position, and depth of objects, both in estimating the depth of the original image and after removing the object.

\begin{table}[bthp]
\centering
\caption{The Effect of Different Depth Priors in Removal NeRF}\label{tab4}
\begin{tabular}{|l|l|c|c|c|c|c|}
\hline
Model & Depth Priors & {PSNR} $\uparrow$ & SSIM $\uparrow$  & {Depth Estimation Time} $\downarrow$ \\
\hline\hline
SpinNeRF& Complete Depth & 21.943 & 0.192  & 44.5s/per image \\
 & ZoeDepth & \textbf{23.451} & \textbf{0.192}  & \textbf{0.58s/per image} \\
 \hline
\end{tabular}
\end{table}

\begin{figure}[htbp]
  \centering
  
  \begin{subfigure}[b]{0.2\textwidth}
    \includegraphics[width=\textwidth]{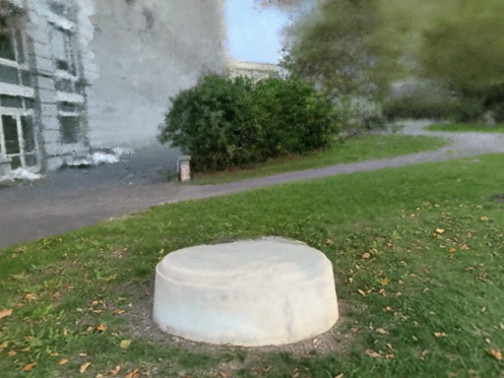}

  \end{subfigure}
  \begin{subfigure}[b]{0.2\textwidth}
    \includegraphics[width=\textwidth]{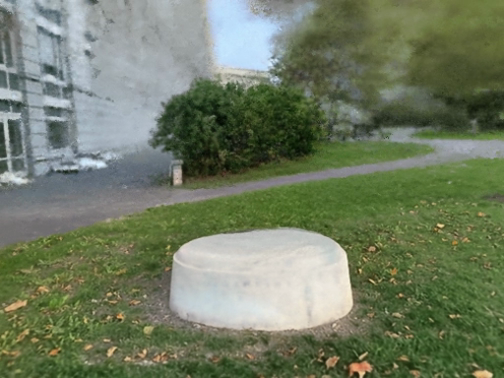}
 
  \end{subfigure}
    \begin{subfigure}[b]{0.2\textwidth}
    \includegraphics[width=\textwidth]{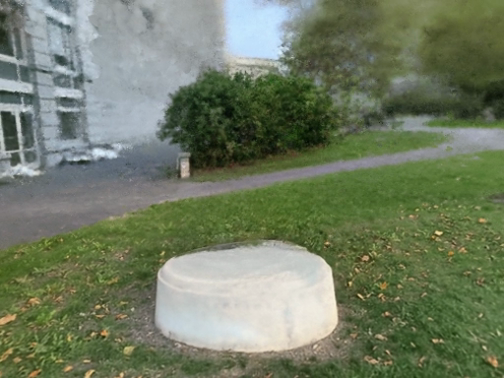}
 
  \end{subfigure}

  \begin{subfigure}[b]{0.2\textwidth}
    \includegraphics[width=\textwidth]{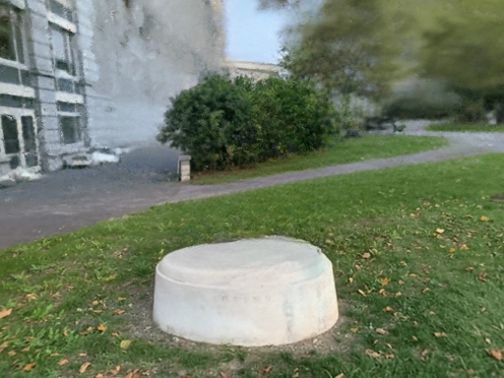}
  \end{subfigure}
    \begin{subfigure}[b]{0.2\textwidth}
    \includegraphics[width=\textwidth]{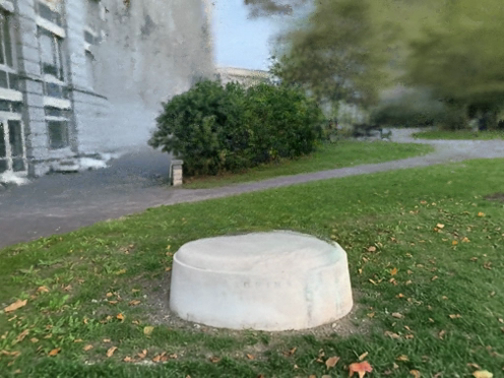}
  \end{subfigure}
  \begin{subfigure}[b]{0.2\textwidth}
    \includegraphics[width=\textwidth]{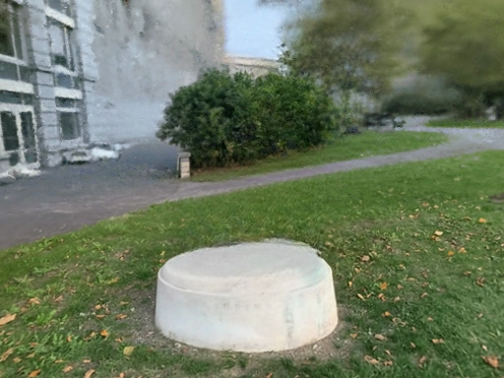}
  \end{subfigure}
  \caption{Rendered views.\textit{Top Row}: depth priors from DSNeRF; \textit{Bottom Row}: depth priors form ZoeDepth.}
  \label{Rendered Views Comparison}
\end{figure} 

\begin{figure}[htbp]
  \centering
  
  \begin{subfigure}[b]{0.3\textwidth}
    \includegraphics[width=\textwidth]{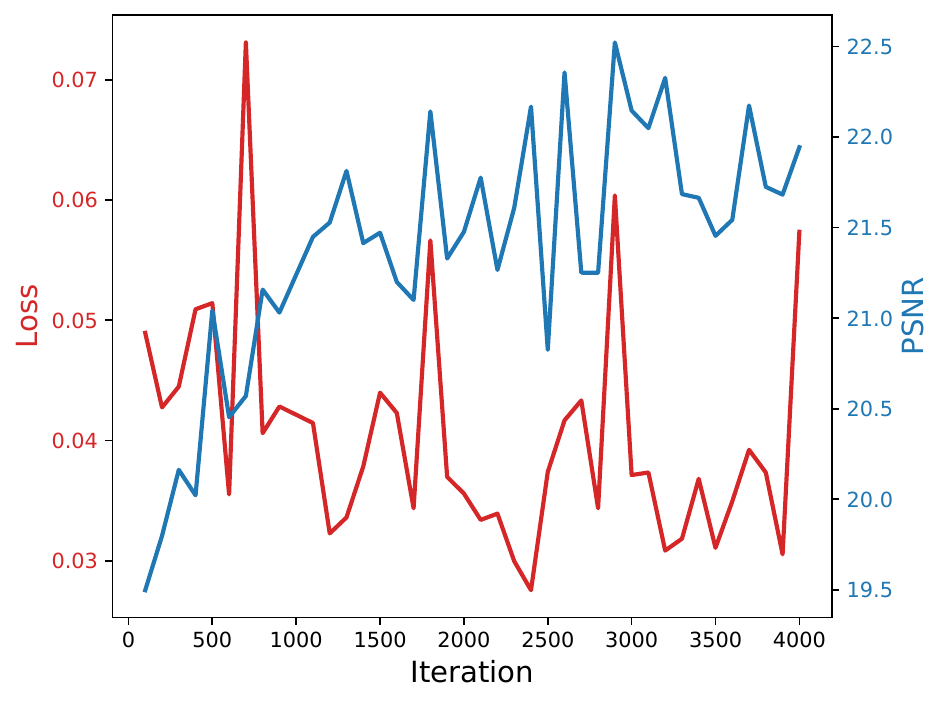}
    \caption{DSNeRF Depth Priors}
  \end{subfigure}
  \begin{subfigure}[b]{0.3\textwidth}
    \includegraphics[width=\textwidth]{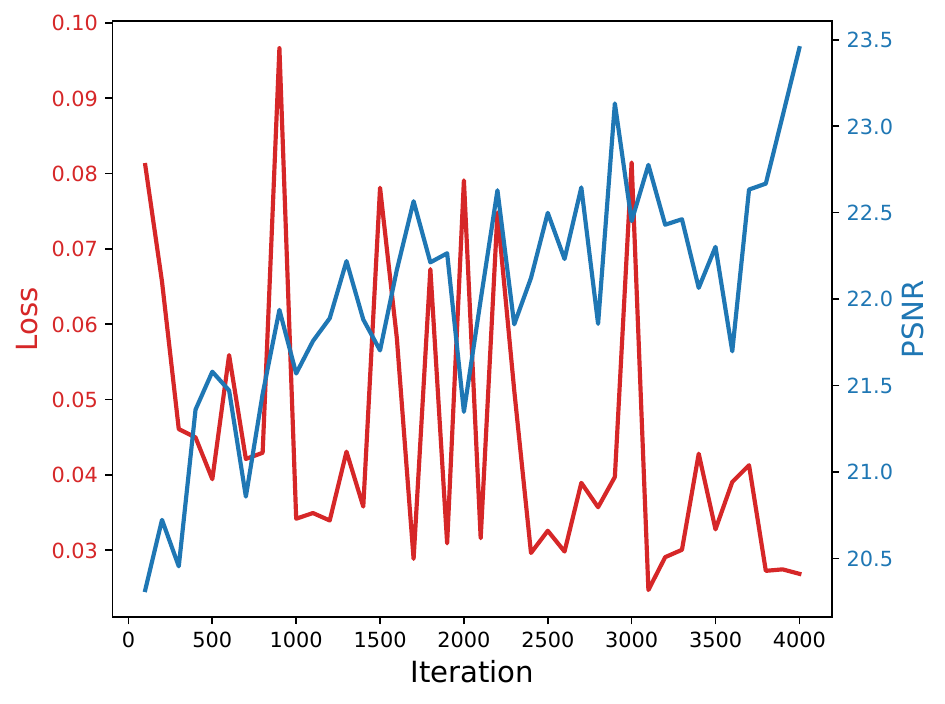}
    \caption{ZoeDepth Priors}
  \end{subfigure}
  \caption{The new pipeline training process with different depth priors.}
  \label{train_process}
\end{figure} 

We analyzed the training dynamics of the proposed pipeline that integrates different depth priors into SpinNeRF for object removal, with details illustrated in Figure \ref{train_process}. This analysis shows a decrease in the model's loss values and an increase in the PSNR values as the number of iterations grows. Specifically, Figure \ref{train_process} (b) demonstrates that incorporating ZoeDepth into SpinNeRF results in a lower and more consistent loss profile and higher PSNR at the same iteration counts compared to using DSNeRF depth priors. A detailed comparison of rendering quality is presented in Figure \ref{Rendered Views Comparison}, where each column represents the same sequence of rendered images. The top row displays the views rendered using DSNeRF depth priors, while the bottom row presents results using ZoeDepth depth priors. Notably, the rendering quality improves significantly with ZoeDepth and depth priors generator, since it helps to retain details such as the walls in the images.


\section{Conclusion}
\label{con}
This paper introduces a new removal NeRF pipeline that innovatively incorporates depth priors, specifically introducing monocular depth estimation models such as ZoeDepth into the SpinNeRF architecture to enhance object removal performance. This approach not only improves SpinNeRF’s capability in handling complex removal scenarios but also significantly reduces computational overhead. Additionally, our extensive evaluation of COLMAP’s dense depth reconstruction on the KITTI dataset positions COLMAP as a cost-effective and scalable alternative to traditional ground truth depth data acquisition methods, a crucial advantage in scenarios constrained by budget limitations. Furthermore, our detailed comparative analysis highlights ZoeDepth as the leading monocular depth estimation method, offering high-quality depth priors with reduced computational demands. The contributions of this paper enhance understanding and further development in NeRF technologies, paving the way for future advancements in digital twin systems and other applications that require robust, detailed 3D reconstructions. 
\bibliographystyle{unsrt}
\bibliography{myp}

\end{document}